\documentclass[11pt,a4paper]{article}
\usepackage{times,latexsym}
\usepackage{url}
\usepackage[T1]{fontenc}

\usepackage{times}
\usepackage{latexsym}
\usepackage{graphicx}
\usepackage{amsmath}
\usepackage{url}
\usepackage{graphicx}
\usepackage{subfigure} 
\usepackage{multirow}
\usepackage{makecell}
\usepackage{colortbl}
\usepackage{amsfonts}
\usepackage{hhline}
\usepackage{tabularx}

\usepackage{amssymb}
\usepackage{booktabs}
\usepackage{colortbl}
\usepackage{makecell}
\usepackage{verbatim}
\usepackage{latexsym}
\usepackage{graphicx,stackengine,scalerel}

\usepackage{stmaryrd}
\usepackage{helvet}
\usepackage{courier}
\usepackage{amsmath}
\usepackage{url}
\usepackage{multirow}
\usepackage{booktabs}
\usepackage{enumitem}
\usepackage{amssymb}
\usepackage{marvosym}
\usepackage{color}
\usepackage{caption}
\usepackage{titlesec}
\usepackage{floatrow}
\usepackage{caption}
\usepackage{rotating}
\usepackage{pifont}
\usepackage{rotating}
\usepackage{xcolor}

\usepackage{colortbl}

\definecolor{shallow_grey}{RGB}{167,194,203}
\definecolor{dark_grey}{RGB}{88,98,103}
\definecolor{light_green}{RGB}{66,195,183}
\definecolor{dark_blue}{RGB}{87,133,149}

\definecolor{dark_red}{RGB}{192,1,0}
\definecolor{dark_orange}{RGB}{255,147,2}
\definecolor{dark_yellow}{RGB}{255,213,121}

\definecolor{card_yellow}{RGB}{255,214,48}
\definecolor{card_blue}{RGB}{167,194,203}
\definecolor{card_green}{RGB}{66,195,183}

\newfloatcommand{capbtabbox}{table}[][\FBwidth]

\usepackage[acceptedWithA]{tacl2021v1}

\usepackage{xspace,mfirstuc,tabulary}

\definecolor{darkgreen}{rgb}{0,0.35,0}

\newcommand{\af}[1]{\textcolor{blue}{#1}}
\newcommand{\tbd}[1]{\textcolor{orange}{#1}}

\newif\iftaclinstructions
\taclinstructionsfalse 
\iftaclinstructions
\renewcommand{\confidential}{}
\renewcommand{\anonsubtext}{(No author info supplied here, for consistency with
TACL-submission anonymization requirements)}
\newcommand{\instr}
\fi

\taclpubformattrue
\iftaclpubformat 

\else

\fi

\title{
Exploring Continual Learning of Compositional Generalization in NLI}

\author{Xiyan Fu \\
  Dept. of Computational Linguistics \\
  Heidelberg University \\
  \texttt{fu@cl.uni-heidelberg.de} \\\And
  Anette Frank \\
  Dept. of Computational Linguistics \\
  Heidelberg University \\
  \texttt{frank@cl.uni-heidelberg.de} \\}

\date{}

\begin{document}
\maketitle
\begin{abstract}
  \textit{Compositional} Natural Language Inference has been 
  explored to assess the true abilities of neural models to perform NLI. Yet,
  current
  evaluations assume models to have full access to all primitive inferences in advance, in contrast to humans that continuously acquire inference knowledge.
  In this paper, we introduce the 
\textit{\underline{C}ontinual \underline{C}ompositional \underline{Gen}eralization in Inference (C$^{2}$Gen NLI)} challenge, where a model continuously acquires knowledge of 
constituting primitive inference tasks as a basis for 
compositional inferences. 
We explore how continual learning affects compositional generalization in NLI, by designing a continual learning setup for compositional NLI inference tasks.
Our experiments demonstrate that models fail to compositionally generalize in a continual scenario. 
To address this problem, we first benchmark various continual learning algorithms and verify their efficacy. 
We then further analyze C$^{2}$Gen, focusing on how to order
primitives and compositional inference types, and examining correlations between subtasks. 
Our analyses 
show that by learning subtasks continuously while observing their dependencies and increasing degrees of difficulty, continual learning can enhance composition generalization ability. \footnote{Data and code can be found in \url{https://github.com/Heidelberg-NLP/C2Gen}.} 

\end{abstract}

\section{Introduction}
Natural Language Inference (NLI) determines the inferential relation between pairs of sentences, by classifying the hypothesis as being true (entailment), undecided (neutral) or false (contradiction) given the premise \citep{dagan2013recognizing, bowman-etal-2015-large, williams-etal-2018-broad}.
The task has been researched for decades and has been shown to facilitate downstream NLU
tasks such as text summarization \citep{laban-etal-2022-summac, utama-etal-2022-falsesum}, question answering \citep{chen-etal-2021-nli-models}, or dialogue generation \citep{stasaski-hearst-2022-semantic}. \par

Recently, large pre-trained models (PLMs) have achieved results on par with human performance by fitting NLI training data \citep{wang2019superglue, raffel2020exploring, chowdhery2023palm}. Despite the success of state-of-the-art PLMs, it remains unclear to what extent neural models have the ability to generalize when performing NLI. 
To better assess the true abilities of PLMs to perform NLI, Compositional Generalization \citep{fodor1988connectionism, hupkes2020compositionality} evaluation has been proposed for NLI \citep{yanaka-etal-2020-neural, geiger-etal-2020-neural, fu-frank-2023-seti}. 
This novel task aims to evaluate whether models are able to predict unseen compositional inferences if they have seen their 
constituting primitive inferences 
in training.
The left part of 
Table \ref{task_comparison} (Compositional Generalization for NLI ) shows 
an unseen compositional NLI test instance for which we expect a model to make the correct prediction
`\textit{He \textcolor{dark_orange}{tries} to \textcolor{light_green}{catch his dog} $\nrightarrow$ He \textcolor{light_green}{catches his pet}}', by relying on the primitive inferences
`\textit{try to S $\nrightarrow$ S}' and `\textit{catch his dog $\rightarrow$ catch his pet}'. 

\begin{table*}[ht]
\centering
\resizebox{\columnwidth}{!}{
\begin{tabular}{llllp{10cm}} \toprule
&\multicolumn{1}{c}{Compositional Generalization for NLI (CGen)}  & & \multicolumn{2}{c}{Continual Compositional Generalization for NLI (C$^2$Gen)}  \\ \hhline{~=~==}
\multirow{5}{*}{\rotatebox{90}{train}} &A man fails to \textcolor{dark_blue}{make a snowball} $\nrightarrow$ A man \textcolor{dark_blue}{plays with a ball}& &\multirow{2}{*}{$\mathcal{S}_{1}$} &A girl \textcolor{dark_orange}{tries} to do a stunt $\nrightarrow$ A girl performs a bicycle trick  \\ 
&A girl \textcolor{dark_orange}{tries} to do a stunt $\nrightarrow$ A girl performs a bicycle trick  &&&A man \textcolor{dark_yellow}{manages} to do a stunt $\nrightarrow$ A man performs a bicycle trick \\ \cmidrule(r){4-5}
&A man fails to \textcolor{light_green}{catch his dog} $\rightarrow$ $\neg$ A man \textcolor{light_green}{catches his pet}&&\multirow{2}{*}{$\mathcal{S}_{2}$}&A man fails to \textcolor{light_green}{catch his dog} $\rightarrow$ $\neg$ A man \textcolor{light_green}{catches his pet} \\
&A man \textcolor{dark_yellow}{manages} to do a stunt $\nrightarrow$ A man performs a bicycle trick && &A man fails to \textcolor{dark_blue}{make a snowball} $\nrightarrow$ A man \textcolor{dark_blue}{plays with a ball} \\ \cmidrule(r){2-2} \cmidrule(r){4-5}
test &A man \textcolor{dark_orange}{tries} to \textcolor{light_green}{catch his dog} $\nrightarrow$ A man \textcolor{light_green}{catches his pet}  & & &A man \textcolor{dark_orange}{tries} to \textcolor{light_green}{catch his dog} $\nrightarrow$ A man \textcolor{light_green}{catches his pet}  \\ \bottomrule
\end{tabular}
}
\caption{
NLI train and test instances for Compositional Generalization in a non-continual (CGen) and continual learning (C$^{2}$Gen) setting. 
Test instances are unseen compositions, while \textcolor{dark_orange}{veridical} and \textcolor{light_green}{NLI} inferences have been seen as primitives during training for compositional inference.
\textit{C$^{2}$Gen} simulates human learning 
via \underline{\textit{a continual learning stream}}, where one primitive task ($\mathcal{S}_{1}$) is learned 
before 
the other 
($\mathcal{S}_{2}$).
In contrast, \textit{CGen} assumes that all data is accessible in advance and \underline{\textit{randomly shuffled}} for training.}
\label{task_comparison}
\end{table*}
                                                 
However, existing works evaluating Compositional Generalization for NLI (\textit{CGen NLI}) rely
on offline training, which crucially differs from the way humans acquire knowledge, i.e., by \textit{continual learning}
\citep{ring1997child, parisi2019continual}. Real communication scenarios require the understanding and induction of compositional inferences relative to dynamically updated knowledge. For example, an agent should be able to compose some newly acquired inferential knowledge 
\textit{\textcolor{dark_blue}{buy an apple vision pro (S)} $\rightarrow$ \textcolor{dark_blue}{digital content is blended with physical space (S')}} 
with previously learned \textit{\textcolor{dark_orange}{try} to S $\nrightarrow$ S}, to induce \textit{\textcolor{dark_orange}{try} to \textcolor{dark_blue}{S} $\nrightarrow$ \textcolor{dark_blue}{S'}}. In Section \ref{sec:application}, we present a promising application of continual compositional inference in a dialogue setting.

To better align 
with the compositional generalization ability in 
real-world situations, and to prepare applying compositional NLI to dynamically evolving information states, we introduce a new
task: \textit{ \underline{C}ontinual \underline{C}ompositional \underline{Gen}eralization for NLI (C$^{2}$Gen NLI)}, which aims to explore
the compositional generalization ability of a model when performing NLI
\textit{in a continual learning scenario}. 
We simulate a continuous learning process by manipulating the order in which specific primitive NLI inferences are encountered during training.
The right part
of Table \ref{task_comparison} shows an example. To solve the unseen compositional inference test sample, a model needs to learn, 
in the first place, the primitive inference \textit{try to S $\nrightarrow$ S} 
($\mathcal{S}_{1}$), and 
then 
\textit{catch his dog $\rightarrow$ catch his pet} ($\mathcal{S}_{2}$).
The C$^{2}$Gen NLI task challenges models in two ways: it tests i) their ability to perform \textit{compositional generalization}, by combining
learned primitive inferences to solve unseen compositional inferences, 
and ii) doing this \textit{in a continual learning scenario} that requires models to memorize and re-use 
primitive inferential knowledge they continually acquired. 
Unlike the existing CGen NLI task, \textit{C$^{2}$Gen NLI} allows us
to evaluate whether
models can learn primitive inferences \textit{continuously} and \textit{efficiently}.

To facilitate research on C$^{2}$Gen NLI, we establish an evaluation setup and task dataset for systematic analysis of the effect that continual learning has on the compositional generalization capabilities of models to perform NLI.
We design two sub-tasks to perform fine-grained compositional generalization analysis: 
i) \textit{compositional inference (Task$_{CI}$}) to explore how well a model performs compositional inference; ii) \textit{primitive recognition (Task$_P$)}, to evaluate a model's ability to resolve
constituting primitive inferences.
With our
evaluation datasets and tasks, we conduct experiments in
CGen and C$^{2}$Gen NLI settings using a multi-task model for the different inference tasks. 
Initial results show that 
with
continual learning, 
models 
show inferior performance in
compositional NLI inference, which we show to be due to \textit{forgetting}.

To combat the forgetting issue, we 
benchmark a set of continual learning algorithms targeted at memorization. Our results validate their
effectiveness, 
but also show 
that memorization alone cannot solve the compositional inference task.
To gain deeper understanding of the challenges 
involved in a continual scenario, we investigate the effect of learning primitive inferences in different orders, analyze correlations between primitive and compositional NLI tasks, and the impact of ordering compositional inference types 
by difficulty. Our findings highlight the importance of ordering inference types in continual learning according to dependencies and intrinsic difficulty.

Our main contributions are as follows:
\begin{enumerate}[label=\roman*), noitemsep]
    \item We motivate and
introduce 
the \textbf{C$^{2}$Gen NLI} (\underline{C}ontinual \underline{C}ompositional \underline{Gen}eralization for \underline{N}atural \underline{L}anguage \underline{I}nference) task, which to our knowledge is the first challenge to explore the \textit{compositional generalization} ability of NLI 
    in a \textit{continual learning scenario}.
    \item We construct a \textit{compositional NLI dataset} and rearrange its partitions for C$^{2}$Gen NLI.
    \item 
    Experiments indicate that \textit{forgetting} is a major challenge for  C$^{2}$Gen NLI. To combat this issue, we benchmark a set of continual learning algorithms and verify their effectiveness.
    \item Further analyses 
    highlight the 
    impact of \textit{guiding the order of continual learning} by observing 
    \textit{dependencies} and \textit{degrees of  difficulty} of primitive and compositional inference types,
    for 
    compositional NLI performance.
    \item By controlling
    for \textit{data leakage} using pseudo data, we 
    demonstrate that the C$^2$Gen NLI challenge persists for LLMs such as Llama. 

\end{enumerate}

\section{Related Work}
Natural language inference (NLI) determines the inferential relation between a hypothesis and a premise \citep{dagan2013recognizing, bowman-etal-2015-large, lai-etal-2017-natural, williams-etal-2018-broad, welleck-etal-2019-dialogue}. Prior works 
aimed to improve
NLI performance with
various neural model types \citep{parikh-etal-2016-decomposable, gong2017natural,chen-etal-2018-neural-natural,bauer-etal-2021-ernie}. Recently, large PLMs
perform well on the NLI task, often
achieving human performance \citep{wang2019superglue, liu2019roberta}. Despite the success of state-of-the-art LLMs, it remains unclear if models are able to generalize when performing NLI.
To better assess their inference abilities, research 
has started to explore
to what extent they 
perform generalization when performing
NLI. This 
includes cross-genre 
\citep{williams-etal-2018-broad} and cross-lingual \citep{conneau-etal-2018-xnli} generalization or investigating the impact of heuristics \citep{mccoy-etal-2019-right, bhargava-etal-2021-generalization}. In this work, we 
evaluate the generalization ability in NLI  
focusing on compositional generalization.
I.e., we test a model's capability of 
predicting unseen compositional inferences if constituting primitive inferences have been learned.

Early works evaluate compositional generalization for NLI targeting
novel compositions involving specific linguistic phenomena, e.g., composing predicate replacements and embedding quantifiers \citep{yanaka-etal-2020-neural}, focusing on lexical entailment and negation \citep{geiger-etal-2020-neural, goodwin-etal-2020-probing}.
Recently, \citet{yanaka-etal-2021-exploring, fu-frank-2023-seti} extended the scope of compositional generalization evaluation 
to composition of veridical inference with customary NLI, finding that PLMs are limited in compositionality. 
Despite promising findings of the above 
works, they all assume that models have full access to all training data in advance.
This is in contrast with humans acquiring knowledge in a continuous fashion.

To simulate human learning processes, continual learning has been proposed \citep{mccloskey1989connectionist, wu2022pretrained}, 
enabling models to learn from a continuous data stream over time.
\citet{robins1995catastrophic, french1999catastrophic} 
identified catastrophic forgetting 
being 
the main challenge in continual learning. To address this issue, 
various
con\-tin\-ual learning strategies 
have been proposed. I.a., data-based algorithms \citep{Chaudhry2019OnTE, chaudhry2019efficient, aguilar2020knowledge} are well-known. They use  small memories to store seen training data, 
to be
reused in later training steps. 
Using such
strategies, later works designed elaborate models to enhance the performance of 
tasks such as relation extraction \citep{wang-etal-2019-sentence}, multilingual learning \citep{berard-2021-continual, mhamdi-etal-2023-cross}, or dialogue 
\citep{madotto-etal-2021-continual}. By contrast, we use such continual strategies to analyze the impact of continual learning on compositional generalization ability in NLI.

Both compositional and continual learning are
pivotal aspects for evaluating the genuine capabilities of large PLMs. Existing works \citep{dziri2023faith, berglund2023reversal, mitchell2023comparing}  indicate that although LLMs are pre-trained on large amounts of data, they still struggle in 
novel tasks and situations. 
Thus, LLMs are expected to learn compositionally and continuously. 
Some recent work aims to combine continual learning and compositionality. 
\citet{li2020compositional} focus on continual learning in a sequence-to-sequence task. They propose to represent syntactic and semantic knowledge separately which allows to leverage compositionality for knowledge transfer.
\citet{jin-etal-2020-visually} introduce a challenging benchmark that aims at continual learning of compositional semantics from visually grounded text. 
Unlike them,
we introduce a new task that focuses on Continual Learning of Compositional Generalization in NLI.
With this task, we i) analyze the challenge of compositional generalization in NLI in a continual learning setup; 
ii) identify the effect of ordering primitive and compositional inference types according to their dependencies and 
difficulty; iii) Finally, in \S\ref{sec:application} we showcase the relevance of continual learning in  NLI in a concrete application, namely, \textit{Persona Dialogue}.

Our finding ii), which highlights the impact of ordering primitive and compositional inference types based on their difficulty, is close to another machine learning paradigm, known as \textit{curriculum learning} \citep{elman1993learning, krueger2009flexible, bengio2009curriculum, soviany2022curriculum}. This learning paradigm is inspired by the human classroom, and refers to training a model with a curriculum of increasing difficulty. Existing works first focus on assessing the difficulty of training samples. According to their difficulty, they further weight data samples and bias the model towards them \citep{kumar2010self, huang-du-2019-self}, or organize data into subgroups and commence learning from the easiest batch \citep{xu-etal-2020-curriculum, jia-etal-2023-sample, ranaldi-etal-2023-modeling}. 
\textit{Curriculum learning} differs from
\textit{continual learning} 
in two respects:\footnote{We refer to Table 2 in \cite{biesialska-etal-2020-continual} for a more comprehensive comparison.} i) \textit{learning schema}. Curriculum learning remains an offline learning method. It focuses on structuring the learning process to facilitate faster and more robust learning, instead, continual learning aims to adapt to new data over time while preserving past knowledge; ii) \textit{training atoms}. Curriculum learning concentrates on data points, instead, continual learning focuses on tasks or knowledge levels. 
Despite these distinctions, curriculum learning and continual learning interact, such as adopting the ordering principle from curriculum learning to enhance continual learning.
Our findings, derived from the analysis of learning sequences in continual learning, could serve as empirical evidence supporting the principles of curriculum learning.



\section{Task Setup: C$^2$Generalization in NLI}
In this section, we provide an overview of continual learning (\S\ref{Continual Learning Bg}) and describe the construction of our Compositional NLI dataset (\S\ref{Compositional Inference}). Building upon this foundation, we rearrange partitions of the dataset to establish \textit{\underline{C}ompositional generalization} tests with standard training (CGen) and a \textit{\underline{C}ontinual learning} (C$^{2}$Gen) setup (\S\ref{Compositional Test}). 

\subsection{Continual Learning Preliminary}
\label{Continual Learning Bg}
Continual learning \citep{mccloskey1989connectionist, wu2022pretrained} is proposed to simulate human learning processes, enabling models to learn from a continuous and non in-distribution data stream over time. The objective is to
enable a model to continuously learn 
a set of instances sequentially ordered with respect to a set of $n$ tasks 
\{${\mathcal{T}_{1},\mathcal{T}_{2}, ..., \mathcal{T}_{n}}$\}, following a given order. The model is trained on examples from $\mathcal{T}_{1}$, progresses to $\mathcal{T}_{2}$, and so on until $\mathcal{T}_{n}$. Notably, during the learning process for each task $\mathcal{T}_{i}$, the model is not allowed to access training data from previous tasks $\mathcal{T}_{<i}$ or future tasks $\mathcal{T}_{>i}$. Within each task $\mathcal{T}_{i}$, instances are trained in a random order. In contrast, conventional training involves full access to all data in advance, meaning the model is trained simultaneously on examples randomly sampled from the set of tasks $\mathcal{T}$.

\subsection{Compositional NLI} 
\label{Compositional Inference}
We model Compositional Inference (CI) building on
customary NLI samples. Both customary and compositional NLI involve the relation between premise and hypothesis, but
com\-po\-si\-tio\-nal inference involves
at least two different primitive inference types (Table \ref{task_comparison}).\footnote{
We restrict ourselves to two primitive 
components.}
To master
compositional inference, a model must 
i) resolve the involved primitive NLI inferences and 
ii) compose the inferred results, using a suitable
composition function. 

We construct compositional inferences by selecting \textit{veridical inference} as a special primitive inference type, and combine it with customary NLI inference samples as a second
primitive inference (cf.\ Table \ref{task_comparison}). Given that veridical inference involves an embedded sentence, it can be flexibly combined and 
scaled to compositional inference datasets  
\citep{yanaka-etal-2021-exploring}. Veridical inference \citep{karttunen1971implicative, ross-pavlick-2019-well} is strongly determined by the lexical meaning of sentence embedding verbs. In the context of a factive veridical verb, we can infer that the proposition it embeds can be
held to be true, e.g., \textit{He manages to S $\rightarrow$ S}. For a non-veridical verb, 
we cannot infer the truth or falsity of a proposition, e.g., \textit{He tries to S $\nrightarrow$ S}; while for non-factive veridical  
verbs, we can infer the negation of the complement,
e.g., \textit{He refuses to S $\rightarrow$ $\neg$ S}. For customary NLI we distinguish three classes:
e(ntailment): S $\rightarrow$ S', n(eutral): S$\nrightarrow$ S', c(ontradiction): S $\rightarrow$ $\neg$S'.

\begin{table}[t]
\centering
\resizebox{0.8\columnwidth}{!}{
\begin{tabular}{@{}cccc@{}} \toprule
index& $P^{V}$& $P^{N}$& $CI$\\ \midrule
\ding{172}& positive& entailment& entailment\\  
\ding{173}& positive& neutral& neutral\\  
\ding{174}& positive& contradiction& contradiction\\ 
\ding{175}& neutral& entailment& neutral\\  
\ding{176}& neutral& neutral& neutral\\  
\ding{177}& neutral& contradiction& neutral\\ 
\ding{178}& negative& entailment& contradiction\\  
\ding{179}& negative& neutral& neutral\\  
\ding{180}& negative& contradiction& entailment\\ 
\bottomrule
\end{tabular}}
\caption{Composition rules for compositional inferences. $P^{V}$, $P^{N}$, and $CI$ indicate 
veridical, customary and compositional inference, respectively.
}
\label{tab_CI_rules}
\end{table}

To construct 
\textit{compositional} NLI samples, we ensure 
that the hypothesis of a (non-)veridical inference pair (\textit{x \textit{verb} S, S})
matches the premise of a customary NLI
pair $(S, S')$, 
to derive a transitive inference pair that may be 
entailed, neutral or contradictory.
E.g, \textit{He tries to do S $\nrightarrow$ S} \& \textit{S $\rightarrow$ S'} $\Rightarrow$ \textit{He tries to do S $\nrightarrow$ S'}. We use the composition rules listed in Table \ref{tab_CI_rules} to define the compositional inference labels. 
\if false
\textcolor{blue}{Alternative:}
To construct valid compositional inferences, we instantiate the complement S of a veridical verb with the premise S$_p$ of an existing NLI pair, and use the NLI pair's hypothesis S$_q$ as the hypothesis of the veridical inference. In this way we perform a transitive inference:\\ \textit{A} \{\textit{starts} $\mid$ \textit{hopes} $\mid$ \textit{refuses}\} \textit{to} $S_p$ \{$\rightarrow \mid \nrightarrow \mid \rightarrow \neg$ \} \textit{ S$_q$}, which the model can resolve by i) computing the veridical inference: \textit{A} \{\textit{starts} $\mid$ \textit{hopes} $\mid$ \textit{refuses}\} \textit{to} $S_p$, ii) $S_p$ \{$\rightarrow \mid \nrightarrow \mid \rightarrow \neg$ \} S$_q$ and iii) apply composition rules on the primitive inference results, which are displayed in  Tab.\ \ref{tab_CI_rules}. 
\fi
For example, \textit{A man \textcolor{dark_orange}{tries} to \textcolor{light_green}{catch his dog} $\nrightarrow$ A man \textcolor{light_green}{catches his pet}} is a (non-entailing) compositional inference. Here, \textit{\textcolor{dark_orange}{tries to S $\nrightarrow$ S}} represents a non-veridical (neutral) 
inference sample, and \textit{\textcolor{light_green}{catch his dog $\rightarrow$ catch his pet}} 
an entailing inference sample. Composing the above primitive inference
results determines the 
label for the compositional inference, 
i.e., \textit{neutral} (rule \ding{175}).

\subsection{Compositional Generalization Testing} 
\label{Compositional Test}
\paragraph{\underline{C}ompositional \underline{Gen}eralization (CGen) in NLI}
Compositional generalization tests are designed to evaluate whether
a model can generalize to unseen compositional inferences whose constituting primitives 
have been observed in training. For example, we can evaluate a model's compositional generalization ability by testing it on an unseen compositional sample \textit{A man \textcolor{dark_orange}{tries} to \textcolor{light_green}{catch his dog} $\nrightarrow$ A man \textcolor{light_green}{catches his pet}}, where its constituting primitive inferences \textit{tries to S $\nrightarrow$ S} and \textit{catch his dog $\rightarrow$ catch his pet} have been seen
in training. We denote the set of possible veridical inference types with $\mathcal{V}$, the set of customary inference types with $\mathcal{N}$, and the set of all possible compositional inference types with $\mathcal{C} = \mathcal{V} \times \mathcal{N}$. 
The domain of all instances of the respective types is given as
$D = \{(v,n) | v\in\mathcal{V}, n\in\mathcal{N}, (v,n)\in\mathcal{C} \}$. 
In all our
compositional generalization experiments, we guarantee there is no intersection between the compositional types used in training and test, i.e., $\mathcal{C}_{train} \cap \mathcal{C}_{test} = \emptyset$, while primitive inferences involved in test instances are ensured to have been seen in training:
$\mathcal{V}_{test} \subseteq \mathcal{V}_{train}$, $\mathcal{N}_{test} \subseteq \mathcal{N}_{train}$. \par

\begin{figure}
    \centering
    \includegraphics[width=2.9in]{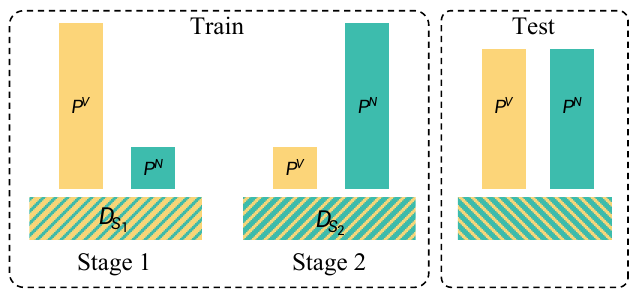}
    \caption{Training and testing setup for compositional inference 
    for continual
    stages $\mathcal{S}_i$, in C$^{2}$Gen. 
    In $\mathcal{S}_1$ we feed various veridicality samples and a few primitive NLI samples.
    $\mathcal{S}_2$ works analogously.
    }
    \label{fig:c2gen_train}
\end{figure}

\paragraph{\underline{C}ontinual \underline{C}ompositional \underline{Gen}eralization (C$^{2}$Gen) in NLI}
Unlike 
standard
compositional generalization evaluation that relies on offline learning, requiring
all training data to be processed in advance, the continual compositional generalization test (C$^{2}$Gen) extends the evaluation to a continual learning setup.
Here, a model 
is fed with
a non-stationary data stream, i.e., the training process follows a controlled learning order, si\-mu\-la\-ting how humans acquire knowledge from their environment. 
Following the standard CGen setup, 
we 
evaluate a model's generalization ability in compositional NLI by testing \textit{unseen} composition types, e.g., \textit{A man \textcolor{dark_orange}{tries} to \textcolor{light_green}{catch his dog} $\nrightarrow$ A man \textcolor{light_green}{catches his pet}}. 
During training, we separate the training stream into \textit{sequential stages} $\mathcal{S}_{i}$ ($i$ $\in$ \{1,2\}), where i) in one stage 
the model learns to categorize \textit{veridical inference} based on the embedding verb 
(e.g., the neutral verb \textcolor{dark_orange}{\textit{try}}); ii) 
in the other it learns to 
categorize a \textit{customary NLI pair} (e.g., the entailment pair
\textit{\textcolor{light_green}{catch his dog $\rightarrow$ catch his pet}}). 
Hence, the model 
first learns one primitive (e.g., $\mathcal{V}$) 
to solve 
compositional inference and then 
the 
other ($\mathcal{N}$), or vice versa. 

We construct the above continual scenario by controlling irrelevant variables. When exploring veridical inference in $\mathcal{S}_{1}$, we 
use a small number of primitive NLI
samples and feed various veridicality samples. Similarly, in $\mathcal{S}_{2}$, we fix a restricted number of
samples from veridical inference and feed various primitive
NLI instances. Parallel to training primitives, compositional instances are presented, 
where the used primitives have been seen in training of the corresponding stage $S_i$.
Different stages are trained sequentially, while samples are randomly trained within each stage. This process enables models to learn incrementally from new data. Fig.\ \ref{fig:c2gen_train} shows the process. 

Compared to customary offline training, 
C$^2$Gen NLI is more challenging and innovative. Because models 
need to learn how to compose primitive inferences, 
and need to preserve previously acquired knowledge of constituting primitive inferences.\par

\section{Analyzing C$^2$Gen NLI as a Multi-Task}

\subsection{Decomposing  Compositional NLI}

\begin{figure}
\centering
\includegraphics[width=76mm]{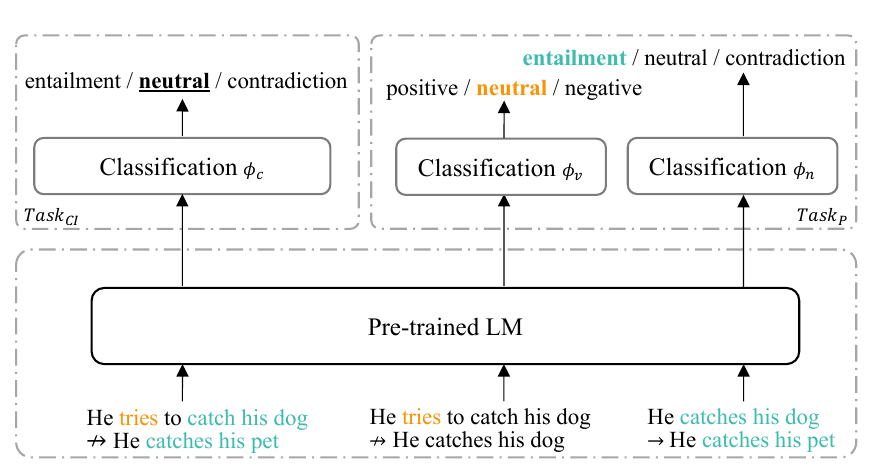}
\caption{Multi-task architecture for compositional generalization evaluation in CGen \& C$^{2}$Gen NLI. Task$_{CI}$ and Task$_{P}$ are jointly optimized.}
\label{fig:model}
\end{figure}

To prepare a deep analysis of the generalization capabilities in C$^2$Gen NLI, i.e., compositional NLI in a continual learning training regime,
we decompose the CGen task into two constituting subtasks: prediction of primitive inferences (Task$_{P}$), and prediction of compositional NLI (Task$_{CI}$) as the main task. 
We apply multi-task learning to jointly learn the two tasks.\footnote{While we expect that task performance will generally profit from MTL with the decomposed subtasks, our main interest is the ability to analyze the effect of continual learning in more detail.} Fig.\ \ref{fig:model} gives an overview.

\paragraph{Task$_{CI}$: \underline{C}ompositional \underline{I}nference} 

In the NLI \textit{CI} task, a model is tasked to predict the inferential relationship instantiated in a given compositional NLI sample. For example, the model is expected to predict the value `\textbf{neutral}' for \textit{A man \textcolor{dark_orange}{tries} to \textcolor{light_green}{catch his dog} $\nrightarrow$ A man \textcolor{light_green}{catches his pet}}.\footnote{Table \ref{tab_CI_rules} shows how the CI NLI value is semantically determined from its constituting NLI primitives.} \par

\paragraph{Task$_{P}$: \underline{P}rimitives Recognition}
Task$_{P}$
evaluates whether a model correctly predicts the 
primitive inferences from which a given compositional sample is built.
I.e., for
\textit{A man \textcolor{dark_orange}{tries} to \textcolor{light_green}{catch his dog} $\nrightarrow$ A man \textcolor{light_green}{catches his pet}} we test the model predictions for its constituting primitive inferences,
expecting i) \textcolor{dark_orange}{neutral} for 
\textit{A man \textcolor{dark_orange}{tries} to S $\nrightarrow$ S} and ii) \textcolor{light_green}{entailment} for 
the entailed inference 
\textit{A man \textcolor{light_green}{catches his dog} $\rightarrow$ A man \textcolor{light_green}{catches his pet}}. 

\subsection{Model}

The \textbf{Compositional Inference (Task$_{CI}$)} is defined as a classification task. The model receives as input the concatenation of the premise and the hypothesis of a compositional NLI sample. The model encodes the sequence to a representation $x$, and we adopt a softmax classifier on top of the classification token of the last 
layer to predict one of the NLI classes, based on the encoded input representation. 
We use the cross entropy function to calculate the compositional reasoning loss $\mathcal{L}_{cr}$
\begin{equation}
    \mathcal{L}_{cr} = \sum_{(x, c)}{} \ell_{CE}(\phi_{c}(x), c)
\end{equation}
where $x$ is the input representation and $c$ the ground truth label. $\phi_{c}$ is the softmax classifier for compositional natural language inference. $\ell_{CE}$ is the cross-entropy function.  

We define the \textbf{Primitive Inferences Recognition task (Task$_P$)}
as a joint classification task, where each classifier
is in charge of a primitive inference. 
For each primitive inference,
we form an input sequence by concatenating premise and hypothesis, and process it in the same way as detailed for the compositional inference task. For the classification, we adopt two softmax classifiers, one for each of the respective tasks. In training we use the cross-entropy function to calculate each primitive's loss. The two primitive losses are jointly considered to train the model for the joint multi-task for primitives recognition $\mathcal{L}_{prim}$
\begin{equation}
    \mathcal{L}_{prim} = \sum_{(x, (v,n))}{} \ell_{CE}(\phi_{v}(x_{v}), v) + \ell_{CE}(\phi_{n}(x_{n}), n)
\end{equation}
where $x_v, x_n$ are the respective input representations, $(v,n)$ the corresponding veridical and NLI instances' ground truth labels. $\phi_{v}$, $\phi_{n}$ are the softmax classifiers for veridical and natural language inference, and 
$\ell_{CE}$ is the cross-entropy function. 

In the end, we use the multi-task training strategy to train two tasks, Task$_{P}$ and Task$_{CI}$. Their objectives $\mathcal{L}_{prim}$ and $\mathcal{L}_{cr}$ are jointly optimized during training, using loss $\mathcal{L} = \mathcal{L}_{prim} + \mathcal{L}_{cr}$.

\subsection{Training Settings} 
\paragraph{Compositional Generalization (CGen).} The standard compositional generalization test in NLI relies on \textit{offline training}, where models have
full access to all training data in advance. This setup serves as an upper-bound baseline for our experiments. All 
training data in CGen is \textit{mixed in a random order}. 
We denote this as $D_{train}$ = $D_{\mathcal{S}_{1}+\mathcal{S}_{2}}$. 

\noindent\paragraph{Continual Compositional Generalization (C$^{2}$Gen).} This new training setup evaluates the compositional generalization capability in NLI in a continual learning scenario. The model is restricted
to follow a \textit{non-stationary data stream}, i.e., all compositional NLI training data is presented in \textit{
a specific order} ($D_{train}$ = $D_{\mathcal{S}_{1}}, D_{\mathcal{S}_{2}}$).

\subsection{Continual Learning Strategies}
In order to deeply 
analyze
the challenges of the C$^{2}$Gen NLI task,
we first
benchmark 
well-known continual learning strategies, designed 
to combat forgetting. 
All methods
introduce a small, fixed-size 
so-called
\textit{episodic memory}. 
It
consists of 
samples selected from a previous learning stage, and is used,
in a next
training stage, in different ways:

\noindent\paragraph{Experience Replay (ER).} \citet{Chaudhry2019OnTE} 
utilize samples from a memory directly \textit{for re-training} in future stages. They distinguish three variants: 
a) \textbf{\textit{ER-res(ervoir)}} applies a 
sampling technique
that ensures that each sample
has an equal chance of being selected; b) \textbf{\textit{ER-buff}} guarantees that the size of the memory 
at each training stage $\mathcal{S}_i$ is the same;
c) \textbf{\textit{ER-mir}} \citep{Aljundi2019online} selects re-training data that is most likely to be forgotten in the next training stage.

\noindent \textbf{Averaged Gradient Episodic Memory (A-GEM).} \citet{chaudhry2019efficient} constrain the di\-rec\-tion of the updated gradient. They cal\-cu\-late the gradient $g'$ of the previous training stage on memory data and 
project the updated gradient to a direction that is close to $g'$.


\noindent\textbf{Knowledge Distillation (KD).} \citet{aguilar2020knowledge} apply memory samples to distill and preserve knowledge learned in previous stages, by minimizing the difference between the output predictions from the previous stage and the current stage over memory data.

\begin{table}[t]
\centering
\resizebox{\columnwidth}{!}{
\begin{tabular}{@{}p{2.1cm}p{7.5cm}@{}} \toprule
\makecell[l]{Signature} & Instantiations\\ \midrule
positive (+)  &  manage, begin, serve, start, dare, use, get, come\\  
neutral ($\circ$)  & hope, wish, expect, try, plan, want, intend, appear \\  
negative ($-$) & forget, fail, refuse, decline, remain \\ \bottomrule
\end{tabular}}
\caption{Instantiation of verbs in different signatures used for constructing veridical inference.}
\label{tab_verbs}
\end{table}

\section{Experimental Setup}
\subsection{Dataset Construction and Verification}
\label{sec:human_annotation}
We construct datasets with instances chosen from established NLI datasets. i) For \textit{primitive veridical inference}, 
we select 21 verbs from the dataset of \citet{ross-pavlick-2019-well}. We restricted our choice to verbs with infinitive complements to ease the construction of compositional samples.
Table \ref{tab_verbs} shows the selected verbs for each class label.
ii) For \textit{primitive customary NLI} we extract 2130 instances (e: 747; n: 693; c: 690)  
from SICK  \citep{marelli-etal-2014-sick}, focusing on instances where the inference
is based on
specific semantic relations including synonymy, hyponymy, active-passive diathesis, etc. For compositional inference, we compose samples from these primitive veridical and customary NLI data points, 
as described in \S \ref{Compositional Inference}. All compositional inferences are categorized into nine groups using the composition rules in Table \ref{tab_CI_rules}. Table \ref{tab_statistics} shows the class distribution.\footnote{Here, we use \{e / n / c\} to denote the veridical inference types, instead of\{positive(p) / neutral(n) / negative(n)\}.} The distribution of target class labels (e:n:c) is roughly 1:2:1. 

As the dataset is automatically constructed from existing datasets,
we perform manual human verification to ensure their validity,
following \citet{keysers2020measuring, liu-etal-2022-challenges, liu2024causes}. 
For cost considerations we restricted manual verification to 200 randomly sampled instances.
Two annotators 
specialized in computational linguistics performed
this task. They 
underwent training 
in practice sessions with direct
feedback before starting the 
annotation process. 
Their task was to annotate the correct class (entailment, neutral, or contradiction) for each premise-hypothesis pair, for all three inference types.
 The 
inter-annotator agreement calculated by 
Cohen's kappa 
was 
0.961, 0.954, and 0.917 for 
the respective inference types.

\begin{table}[t]
\centering
\resizebox{\columnwidth}{!}{
\begin{tabular}{ll|ll|ll} \toprule
 type & \#num & type & \#num & type & \#num \\ \midrule
 \ding{172}ee\_e  & 5976  &\ding{175}ne\_n    & 5976  & \ding{178}ce\_c  & 3735 \\
 \ding{173}en\_n  & 5544  &\ding{176}nn\_n   & 5544  & \ding{179}cn\_n   & 3465 \\
 \ding{174}ec\_c  & 5520  &\ding{177}nc\_n   & 5520  & \ding{180}cc\_e & 3450 \\ \bottomrule
\end{tabular}}
\caption{Distribution of the nine compositional inference types in testing. `xy\_z' specifies the values of the respective primitive inferences types, where `x' stands for veridical, `y' for customary NLI, `z' for compositional inference, with `x', `y', `z' $\in$ \{entailment(e) / neutral(n) / contradiction(c)\}\footnotemark[4].}
\label{tab_statistics}
\end{table}

After the annotation,
we computed the consistency between the human-labeled and automatically constructed data for each inference type.
Among incorrect \textit{veridical inference} samples (15 cases)\footnote{We provide the aggregate count of incorrect samples annotated by two annotators for analysis.}, 87\% of instances 
are susceptible of 
a systematic veridicality bias among humans \citep{ross-pavlick-2019-well}. I.e., some verbs with neutral signature are often perceived to have positive signature, while our construction follows the semantic definition (cf.\ Table \ref{tab_CI_rules}). 
The remaining 13\% are due to a range of different annotation errors.
For \textit{customary inference} (based on SICK), we follow the taxonomy of \citet{kalouli2023curing} to categorize error samples (24 cases). Applied to our data, the errors are attributed to the following sources:
ambiguous (55\%), looseness (25\%), phrasal verbs (10\%), and annotation error (10\%).
Note that NLI labeling consistency, in general, 
is still an open issue, relating to factors such as
ambiguity 
and uncertainty \citep{pavlick-kwiatkowski-2019-inherent,nie-etal-2020-learn, jiang-marneffe-2022-investigating}. For incorrect \textit{compositional inferences} (25 instances), we note that incorrect primitive inferences cause accumulated errors, accounting for approx.\ 91.5\% of the incorrect compositional inferences. The remaining ones are annotation errors.
Still, the 
consistency 
for each
inference type exceeds 90\%, indicating a high quality of our benchmark dataset, which can be a valuable resource for future work.

\subsection{Dataset Split}
The compositional inference data $\mathcal{D}^{C}$ is prepared 
for the \textbf{Compositional Generalization in NLI (CGen)} evaluation as follows:
Given nine compositional inference types, we conduct nine-fold cross-validation experiments, reporting averaged results. Specifically, each type will once serve as a test dataset $\mathcal{D}_{test}^{C}$ (e.g., \ding{172}), while the remaining eight types are used as training set $\mathcal{D}_{train}^{C}$ (e.g., \ding{173}-\ding{180}).
As outlined in \S \ref{Compositional Test}, we guarantee that the primitive inferences used in a given test instance have been seen in training.  
For Task$_{CI}$ we train on $\mathcal{D}_{train}^{C}$ and test on $\mathcal{D}_{test}^{C}$
. For Task$_{P}$ we decompose the instances of $\mathcal{D}_{train}^{C}$ and $\mathcal{D}_{test}^{C}$ into their primitive inferences $\mathcal{D}_{train}^{C}$ $\Rightarrow$ $\mathcal{D}_{train}^{P_{V}}$ \& $\mathcal{D}_{train}^{P_{N}}$ for primitive recognition training, and test with unseen primitive inferences $\mathcal{D}_{unseen}^{P_{V}}$, $\mathcal{D}_{unseen}^{P_{N}}$.

In the \textbf{Continual Compositional Generalization in NLI (C$^2$Gen)} setting (cf.\ \S \ref{Compositional Test}) we maintain the evaluation protocol for both tasks as detailed above for CGen, but split the train set $\mathcal{D}_{train}^{C}$ into $\mathcal{D}_{\mathcal{S}_{1}}^{C}$ and  $\mathcal{D}_{\mathcal{S}_{2}}^{C}$ s.th. $\mathcal{D}_{train}^{C}$ = \{$\mathcal{D}_{\mathcal{S}_{1}}^{C}$ $\cup$ $\mathcal{D}_{\mathcal{S}_{2}}^{C}$\}, and present this data in a continual training stream.
For each stage $\mathcal{S}_{i}$, $i \in$ \{1,2\}: i) If it serves to train the model to learn \textit{veridical inference}, we use a small number of NLI samples and feed various veridicality samples. E.g., we select data from \ding{173}\ding{176}\ding{174}\ding{180}, where for the pair \ding{173}\ding{176} the model needs to distinguish the effect of positive and neutral veridicality and similarly for \ding{174}\ding{180}, where it needs to distinguish the effect of positive and negative veridicality. ii) If the model is tasked to learn \textit{natural language inference}, we use a small number of veridical verbs, selecting data from \ding{175}\ding{177}\ding{178}\ding{179} (for similar reasons as in i). We experiment with alternative data streams, with reversed order in which specific phenomena are trained, once setting $\mathcal{S}_{1}$ to process training data 
targeted to $\mathcal{V}$ and $\mathcal{S}_{2}$ to $\mathcal{N}$, and once choose the opposite assignment to $\mathcal{S}_{1}$ and $\mathcal{S}_{2}$. In each stage, we uniformly sample 3200 instances for training.

\subsection{Evaluation Metric}
We adopt two metrics: 
i) \textit{Acc(uracy)} reports the percentage of correctly predicted labels for a given task after training on all 
stages. 
ii) \textit{Forget} is a commonly used metric in
continual learning. 
It measures to what extent knowledge that a model has learned in $\mathcal{S}_{1}$ 
is preserved after training in $\mathcal{S}_{2}$.
For a given task $T$, \textit{Forget} is calculated as ($Acc_{\mathcal{S}_{1}}(D_{test}^{T}) - Acc_{\mathcal{S}_{1}, \mathcal{S}_{2}}(D_{test}^{T})$) / $Acc_{\mathcal{S}_{1}}(D_{test}^{T})$.

\subsection{Implementation Details}
\if false
In this section, we outline implementation details 
for our subsequent evaluations and analysis. 
\fi 
\paragraph{Backbone.} 
We use RoBERTa-Large\footnote{\url{https://huggingface.co/roberta-large}} \citep{liu2019roberta} 
 given its superior NLI performance\footnote{\url{https://gluebenchmark.com/leaderboard}} and training efficiency, following \citep{yanaka-etal-2021-exploring, fu-frank-2023-seti}. We train using Adam Optimizer with a learning rate 1e-5 and batch size 8. 

\paragraph{Continual Learning Strategies.} For all evaluations using continual learning strategies, we set the memory size to 100.
Following \citet{Chaudhry2019OnTE, Aljundi2019online}, 
we set the number of replay samples in each step to the batch size for ER-based strategies, including \textit{ER-reservoir}, \textit{ER-buff}, and \textit{ER-mir}. 
In practice, we add the memory batch to the current batch in training.
For fair comparison to other strategies, we set the sample size to be equal 
to the batch size used for controlling the gradient in AGEM \citep{chaudhry2019efficient} and for distilling knowledge in KD \citep{aguilar2020knowledge}. For each experiment, we perform three runs with different seeds, as in \citet{jin-etal-2020-visually, madotto-etal-2021-continual}. We report the mean performance with standard deviations in the following experiments.

\paragraph{Hyperparameter settings.
} 
We determine suitable hyperparameters by empirical assessment in a grid search.
To assess the impact of the learning rate, we run experiments across
a range of learning rates [1e-5, 2e-5, 3e-5] using Adam optimizer.\footnote{We follow \citet{liu2019roberta} in the selection of potential learning rates, as excessively large or small values can impede convergence in RoBERTa.} Results indicate that the gap ($\Delta$) between CGen and C$^{2}$Gen increases monotonically with increasing learning rate, achieving accuracies of 
[18.11, 19.05, 19.88] for Task$_{P}$ and [7.44, 8.39, 9.26] for Task$_{CI}$ for the respective choices.
We select 
1e-5 as the learning rate because its gap is the most negligible compared to the other rates. Moreover, the similarity in gap values between Task$_{P}$ and Task$_{CI}$ implies that adjusting hyperparameters alone does not significantly impact the subsequent conclusions. We similarly evaluate the impact of memory capacity on continual strategies, 
for ranges from 2\% to 5\% of the one-stage training data, corresponding to memory sizes of 50, 100, 150, and 200. Again, the results for the two tasks exhibit a unimodal distribution, with a peak occurring at 100. Therefore, we opt to utilize a memory size of 100.

\begin{table}[t]
\centering
\resizebox{\columnwidth}{!}{
\begin{tabular}{@{}lccccc@{}} \toprule
\multirow{2}{*}{Settings} & \multicolumn{3}{c}{Task$_P$} & \multirow{2}{*}{Task$_{CI}$} \\ \cmidrule(r){2-4}
&V&N&V+N& \\ \midrule  
CGen &99.96\textsubscript{0.12} &94.36\textsubscript{0.57} &94.36\textsubscript{0.41} &46.67\textsubscript{0.26} \\ \midrule
\textit{ver} $\rightarrow$ \textit{nat}  &  & & \\
C$^{2}$Gen ($\mathcal{S}_{1}$) &100.00\textsubscript{0.00} &- &-  &- \\ 
C$^{2}$Gen ($\mathcal{S}_{2}$) &80.72\textsubscript{0.39} &94.25\textsubscript{0.76} & 76.31\textsubscript{0.59} &39.40\textsubscript{0.43} \\ \midrule
\textit{nat} $\rightarrow$ \textit{ver}  & & & \\
C$^{2}$Gen ($\mathcal{S}_{1}$) &- &93.94\textsubscript{0.65} &- &- \\ 
C$^{2}$Gen ($\mathcal{S}_{2}$) &99.58\textsubscript{0.14} &71.15\textsubscript{0.72} &70.73\textsubscript{0.49} &37.36\textsubscript{0.57} \\ \bottomrule
\end{tabular}}
\caption{Results for Task$_P$ (incl. individual primitives) and Task$_{CI}$ in different training settings. Subscripts are the standard deviation.}
\label{tab_q1}
\end{table}

\begin{table*}[ht]
\centering
\resizebox{\columnwidth}{!}{
\begin{tabular}{@{}lcccccccccc@{}} \toprule
\multirow{4}{*}{Settings} & \multicolumn{5}{c}{\textit{ver} $\rightarrow$ \textit{nat}} & \multicolumn{5}{c}{\textit{nat} $\rightarrow$ \textit{ver}} \\  \cmidrule(r){2-6} \cmidrule(r){7-11}
& \multicolumn{4}{c}{Task$_{P}$} & Task$_{CI}$ & \multicolumn{4}{c}{Task$_{P}$} & Task$_{CI}$ \\ \cmidrule(r){2-5} \cmidrule(r){7-10}
&Acc$_{V}$($\uparrow$) &Acc$_{N}$($\uparrow$) &Acc$_{V+N}$($\uparrow$) &Forget$_{V}$($\downarrow$)  &ACC($\uparrow$) &Acc$_{V}$($\uparrow$) &Acc$_{N}$($\uparrow$) &Acc$_{V+N}$($\uparrow$) &Forget$_{N}$($\downarrow$)  &ACC($\uparrow$)  \\ \midrule
C$^{2}$Gen ($\mathcal{S}_{2}$) &80.72\textsubscript{0.39} &94.25\textsubscript{0.76} & 76.31\textsubscript{0.59}  &19.18\textsubscript{0.39} &39.40\textsubscript{0.43} &99.58\textsubscript{0.14} &71.15\textsubscript{0.72} &70.73\textsubscript{0.49} &24.26\textsubscript{0.48}  &37.36\textsubscript{0.57} \\\midrule
ER - Res &99.89\textsubscript{0.01} &94.14\textsubscript{0.56} &94.04 \textsubscript{0.53} &0.11\textsubscript{0.01} &44.89 \textsubscript{0.68} &100.00\textsubscript{0.00} & 87.43 \textsubscript{0.67} &87.43\textsubscript{0.67} &7.64\textsubscript{0.53}  &42.34 \textsubscript{0.71} \\ 
ER - Buff &99.78\textsubscript{0.01} &94.25\textsubscript{0.34} &93.91\textsubscript{0.28} &0.15\textsubscript{0.01} &44.34\textsubscript{0.56} &100.00\textsubscript{0.00} &87.38\textsubscript{0.59} &87.38\textsubscript{0.59} &6.91\textsubscript{0.42}  &41.68\textsubscript{0.42}  \\ 
ER - Mir &99.92\textsubscript{0.00}&94.87\textsubscript{0.29} &94.04\textsubscript{0.19}  &0.08\textsubscript{0.00} &44.73\textsubscript{0.72} &100.00\textsubscript{0.00} &87.55\textsubscript{0.71} &87.55\textsubscript{0.71} &6.71\textsubscript{0.69}  &42.01\textsubscript{0.66}  \\ 
AGEM &99.86\textsubscript{0.02} &94.91\textsubscript{0.87} &93.78\textsubscript{0.75} &0.14\textsubscript{0.03} &42.10 \textsubscript{0.91}&99.61\textsubscript{0.03} &81.70\textsubscript{1.12} &81.35\textsubscript{0.94} &13.25\textsubscript{0.86}  &41.60\textsubscript{0.81}  \\ 
KD &99.80\textsubscript{0.03} &94.56\textsubscript{0.63} &90.13\textsubscript{0.44} & 0.20\textsubscript{0.03} &42.37\textsubscript{0.77} &97.86\textsubscript{0.04} &82.69\textsubscript{0.99} &81.90\textsubscript{0.87} &11.57\textsubscript{0.68}  &41.78\textsubscript{0.74}  \\ 
\bottomrule
\end{tabular}}
\caption{Results of compositional primitive recognition (Task$_P$) and inference (Task$_{CI}$) in C$^{2}$Gen NLI across different continual learning strategies. Subscripts are the standard deviation.} 
\label{tab_strategy}
\end{table*}

\section{Results and Analysis}\label{sec:results}
\subsection{\textbf{How does a model perform in C$^{2}$Gen?}} \par
We start by analyzing the effects of the different training settings, \textit{CGen} and \textit{C$^{2}$Gen}, on model performance in the compositional generalization test for NLI (Task$_{CI}$).
Table \ref{tab_q1} shows the results. In the CGen setting, the model shows decent
performance in compositional inference (Task$_{CI}$) with
an accuracy of 46.67. Compared to CGen, C$^{2}$Gen NLI shows a decline for both continual order variants \textit{ver} $\rightarrow$ \textit{nat} and \textit{nat} $\rightarrow$ \textit{ver}, with reductions of 7.27 and 9.31 points, respectively. 
This suggests that 
compositional generalization in NLI in a continual learning scenario is more challenging.

\textbf{Why is C$^{2}$Gen more challenging?} To investigate this question, we examine the accuracy of 
primitive inference (Task$_{P}$) in different continual learning stages. 
This is because Task$_{CI}$ is dependent on Task$_{P}$, requiring correct predictions for the constituting elements of the composition.
For C$^{2}$Gen in order \textit{ver} $\rightarrow$ \textit{nat}, we find that the initially learned veridical primitive  inference 
achieves high accuracy of 100\% in stage $\mathcal{S}_{1}$, showing
that the model has achieved perfect knowledge of
veridical inference after $\mathcal{S}_{1}$. However, the accuracy for veridicality drops to 80.72 ($\downarrow$19.18) after learning primitive NLI in $\mathcal{S}_{2}$. This suggests that the model forgets the primitive knowledge learned during $\mathcal{S}_{1}$. We find a similar trend in the C$^{2}$Gen setting \textit{nat} $\rightarrow$ \textit{ver}, where the accuracy of the initially learned NLI primitive inference drops from 93.94 to 71.15 ($\downarrow$22.79). While in each order only one primitive is affected by forgetting, the joint accuracy for Task$_P$
drops to 70-76 points in both settings.
From these observations 
we conclude that \textbf{catastrophic forgetting is a major challenge in C$^{2}$Gen}.

\subsection{\textbf{
Can continual learning strategies help?}
}\par
\label{strategy}

Next,  
we 
apply existing continual learning strategies that are designed to address the problem of forgetting, 
and analyze their effect on the preservation of knowledge of primitives (Task$_P$) and on compositional generalization (Task$_{CI}$) in C$^{2}$Gen, for both learning orders. Table \ref{tab_strategy} shows the results. Compared to vanilla 
C$^{2}$Gen,
all continual strategies yield improved accuracy for both 
tasks and
reduce the forgetting value of learned primitive inference. 
C$^{2}$Gen$_{ver \rightarrow nat}$, 
yields a significant improvement in the accuracy of the initially learned primitive (Acc$_{V}$), with an increase from approx.\ 80 to 100. Accordingly,
the forgetting value associated with this primitive 
decreases by the same amount to almost 0.
A similar trend is seen
in C$^{2}$Gen$_{nat \rightarrow ver}$ 
where the accuracy of the initially learned primitive (Acc$_{N}$) increases from 71 to 83, 
while its forget value drops from 24 to 10.
This shows that
\textbf{
continual learning 
strategies alleviate forgetting, helping the model to regain 
substantive
performance} 
(+5 points for Task$_{CI}$).

We then analyze the effect of
different continual strategies. 
Table \ref{tab_strategy} shows that
Experience Replay strategies (ER-Res/Buff/Mir) achieve superior results 
with two tasks in different learning orders. For example, in 
C$^2$Gen$_{ver \rightarrow nat}$ 
ER-based strategies achieve a Task$_{CI}$ accuracy of 44 (as opposed to 42 for AGEM and KD).  
With the reverse order C$^2$Gen$_{nat \rightarrow ver}$ the performance is lower for both tasks: Task$_{CI}$ achieves 42 (ER) vs. 41 (non-ER); Task$_{P}$ yields 87 (ER) vs. 81 (non-ER).  
The only exception is Task$_{P_V}$ in C$^{2}Gen_{ver \rightarrow nat}$, where all continual strategies show comparable performance, 
at almost 100\%. This is likely due to the ease of learning highly lexicalized veridicality classes, to which continual strategies cannot contribute much (cf.\ also Table \ref{tab_q1}).

\section{
Establishing learning order for C$^2$Gen}\label{sec:analysis}
As shown in \S \ref{strategy}, continual 
strategies can greatly improve the performance of primitive and compositional NLI 
in C$^2$Gen NLI. However, the continual learning results still lag behind non-continual training.
To gain deeper understanding of the challenges involved in the continual learning process for compositional generalization inference, we perform further analysis of the C$^2$Gen setting.\footnote{In
this section we select ER-Res as continual learning strategy for our experiments, given its superior performance (cf.\ Table \ref{tab_statistics}). The remaining strategies show similar trends.}

\subsection{Effects of primitive learning orders}
\label{compos_orders}

While it seems evident that primitive tasks must be learned prior to 
compositional tasks they are constitutive for, the order among 
primitive tasks is more difficult to establish. To explore how different orders of learning 
primitives 
in continual learning
affect 
compositional generalization,
we compare the performance of Tasks $P$ and $CI$ with alternating orders of learning \textit{veridical inference} (\textit{ver}) and \textit{customary NLI inference} (\textit{nat}), i.e., \textit{ver} $\rightarrow$ \textit{nat} vs.\ \textit{nat} $\rightarrow$ \textit{ver}. Table \ref{tab_strategy} shows 
that 
\textit{ver $\rightarrow$ nat} consistently outperforms \textit{nat $\rightarrow$ ver}. 
For \textit{ER-Res}, e.g., i) for
Task$_P$, Acc$_{V+N}$ differs by 
6.61 points (94.04 vs.\ 87.43); ii) for Acc$_{CI}$ in Task$_{CI}$ the difference is smaller, but still considerable (2.55 points).
These differences indicate that \textbf{the order of learning constituting primitive inferences is relevant for compositional NLI inferences}. 

\begin{figure}
\centering
\includegraphics[width=1\linewidth]{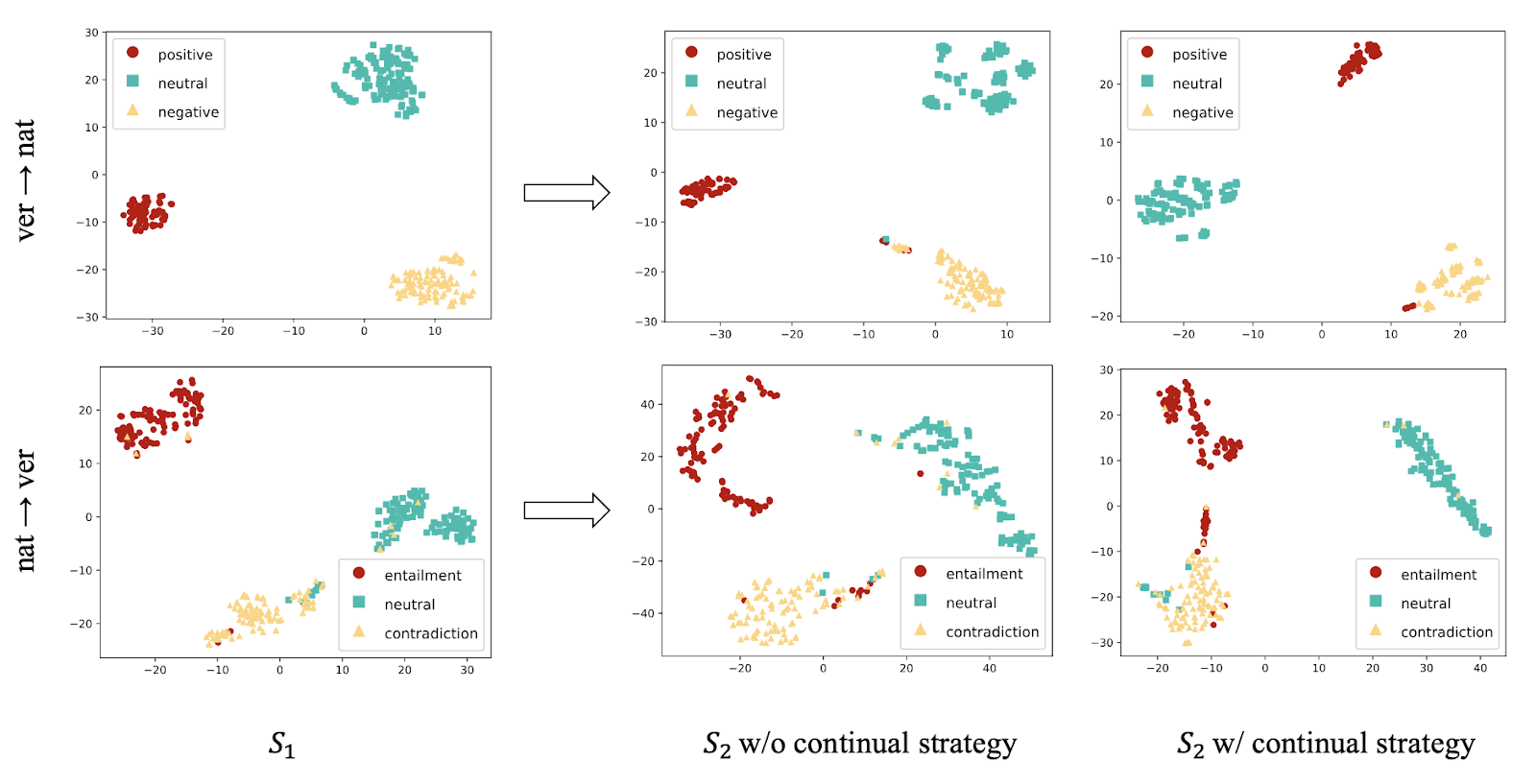}
\caption{Changes of 
learned primitive representations from $\mathcal{S}_1$ to  $\mathcal{S}_2$ 
with different
learning orders.}
\label{fig:order-visual}
\end{figure}

In order to investigate why 
\textit{ver $\rightarrow$ nat} performs better
than \textit{nat $\rightarrow$ ver}, we examine the representation changes of the initially learned primitives for the respective learning orders at different timesteps:
i) by the end of $\mathcal{S}_{1}$, where the model has just completed learning the initial primitive, and ii) after $\mathcal{S}_{2}$, when the model has completed learning of the other primitive. For $\mathcal{S}_{2}$ we compare two settings: pure continual learning ($\mathcal{S}_{2}$ w/o continual strategy) and continual learning using the ER-Res strategy ($\mathcal{S}_{2}$ w/ continual strategy).

Fig.\ \ref{fig:order-visual} visualizes the results. 
For both orders, we observe similar changes between $\mathcal{S}_{1}$ and $\mathcal{S}_{2}$: The three categories within each primitive inference type are clearly grouped in $\mathcal{S}_{1}$. In  $\mathcal{S}_{2}$, the shapes of the three clusters get looser in $\mathcal{S}_{2}$ w/o continual strategy, while with 
continual strategy in $\mathcal{S}_{2}$ (rightmost images), the density of each cluster can be recovered.
When comparing the density of the individual clusters for 
the different orders (\textit{ver $\rightarrow$ nat} vs.\ \textit{nat $\rightarrow$ ver}), it becomes evident that the clusters in \textit{ver $\rightarrow$ nat} exhibit a higher level of density in both stages. This suggests that veridical inference is easier to learn than customary NLI,
leading to reduced likelihood of forgetting. This finding highlights the importance of considering the inherent difficulty of learning a primitive, and to \textbf{order primitives that are easier to learn first}.

\subsection{Continual learning 
of 
dependent 
tasks 
} 
\label{compos_types}

To better understand the challenges of 
compositional NLI 
in the different learning frameworks, we further analyze the correlation between Task$_{P}$ and Task$_{CI}$. 
We decompose the compositional inference testing data into its primitive inferences $\mathcal{D}_{test}^{C}$ $\Rightarrow$ $\mathcal{D}_{test}^{P_{V}}$ \& $\mathcal{D}_{test}^{P_{N}}$ for primitive recognition. We then categorize all test instances into four groups: i) P(\ding{51})CI(\ding{51}), where both tasks yield \textit{correct} predictions. ii) P(\ding{51})CI(\ding{55}), where seen primitive inferences are \textit{correctly classified,} 
but predicting
unseen compositions fails. This 
we identify
as \textit{lacking generalization capability}. iii) P(\ding{55})CI(\ding{51}) records unseen compositions that are correctly predicted without accurately recognizing 
their primitives. Given that Task$_{P}$ is a
prerequisite for Task$_{CI}$, this scenario 
indicates \textit{a shortcut}. iv) for P(\ding{55})CI(\ding{55}), where both tasks are incorrectly predicted,
the model \textit{fails the complete task}. 

\begin{table}[t]
\centering
\resizebox{\columnwidth}{!}{
\begin{tabular}{@{}lcccc@{}} \toprule
Setting &P(\ding{51})CI(\ding{51}) &P(\ding{51})CI(\ding{55}) &P(\ding{55})CI(\ding{51}) &P(\ding{55})CI(\ding{55}) \\ 
Indicates:& \textit{correct} & \textit{no generalization} & \textit{shortcut} & \textit{wrong}\\
\midrule  
CGen &46.05 &52.41 &0.62 &0.92 \\ 
C$^{2}$Gen &37.98(\footnotesize $\Delta$8.07)&46.71(\footnotesize $\Delta$5.70) &1.42(\footnotesize $\Delta$0.80 )&13.89(\footnotesize $\Delta$12.97) \\ 
ER-Res &44.33(\footnotesize $\Delta$1.72) &54.38(\footnotesize $\Delta$1.97) &0.56(\footnotesize $\Delta$0.06) &0.73(\footnotesize $\Delta$0.19) \\ \bottomrule
\end{tabular}}
\caption{Distribution of class performance across Task$_{P\times CI}$ for different settings (all: \textit{ver $\rightarrow$ nat}). $\Delta$ indicates the gap compared to CGen.}
\label{tab_four_cat}
\end{table}

Table \ref{tab_four_cat} displays the distribution of these 
cases.
For CGen, we find
an exceedingly low percentage of instances in the P(\ding{55})CI(\ding{51}) category, indicating a scarcity of model shortcuts.
Since P(\ding{51})CI(\ding{51}) and P(\ding{51})CI(\ding{55}) jointly cover the remaining probability mass, we conclude that the model meets the preconditions for solving Task$_{CI}$ by being able to solve Task$_P$. But, about half of these cases fail to perform compositional NLI inference in Task$_{CI}$. This suggests that evaluated models lack compositionality. In contrast, human annotation evaluations (\S\ref{sec:human_annotation}) show that incorrect compositional inferences mainly stem from accumulated errors in primitive inferences. That is, P(\ding{55})CI(\ding{55}) is more predominant compared to P(\ding{51})CI(\ding{55}). This indicates that humans show greater proficiency in handling compositionality compared to models.

Continual learning in C$^{2}$Gen shows a reduction in the proportions 
of P(\ding{51})CI(\ding{51}) and P(\ding{51})CI(\ding{55}), with the majority of erroneous predictions transitioning to 
P(\ding{55})CI(\ding{55}).
This shows that con\-tin\-ual learning has a clear impact on primitive recognition, with or without generalization ability.
Enhancing the model with 
strategy ER-Res 
yields a reduction for 
P(\ding{55})CI(\ding{55}) 
and a corresponding increase of the 
P(\ding{51})CI(\ding{51}) and P(\ding{51})CI(\ding{55}) classes. However, the increase is more pronounced for the \textit{no generalization class} (+3.7).
I.e., ER-Res proves more effective for
primitives compared to compositional generalization.   
This may be due to the complexity of two tasks, making it relatively easier for primitives to recover from forgetting.
Overall, 
we show that 
memorization methods can alleviate 
the forgetting effect for primitives, while compositional inference remains challenging, with a small decrease compared to CGen.

\subsection{C$^2$Gen by increasing difficulty of  tasks}
As the above analysis 
shows, C$^2$Gen remains challenging with a gap of $\Delta$1.78 for $CI_{ver \rightarrow nat}^{ER-Res}$ compared to CGen (Table \ref{tab_strategy}).
We aim to explore how to relieve this issue. Inspired from our insights into ordering effects for primitive inference types (\S \ref{compos_orders}) and the curriculum learning paradigm, we investigate the effect of ordering the continual learning stream for the complete compositional task along the degree of difficulty for all involved NLI types. 

\begin{table}[t]
\centering
\resizebox{\columnwidth}{!}{
\begin{tabular}{@{}c|ccc|c|l@{}} \toprule
 &N$_e$ &N$_n$ &N$_c$ & Avg.$CI_{V}$ &Function Types \\ \midrule
V$_e$ &\cellcolor{light_green}19.50 &\cellcolor{shallow_grey}73.86 &\cellcolor{dark_yellow}13.91 &35.76 &f$_{v_e}$($v_e$,X) = X\\
V$_n$ &\cellcolor{shallow_grey}100 &\cellcolor{shallow_grey}100 &\cellcolor{shallow_grey}57.21 &85.74 &f$_{v_n}$($v_n$,\_) = $n$\\
V$_c$ &\cellcolor{dark_yellow}13.99 &\cellcolor{shallow_grey}26.50 &\cellcolor{light_green}15.03 &18.51 &f$_{v_c}$($v_c$,X) = $\neg$ X \\ \midrule
Avg. $CI_N$ &44.50 &66.79 &28.72 &46.67 &  \\  \bottomrule
\end{tabular}
}
\caption{Task$_{CI}$ accuracy (CGen) for all inference types. V, N denote
veridical and customary inference. Avg.\ states
average results for different function types. Color indicates the CI target labels:
\raisebox{0.2\baselineskip}{\colorbox{light_green}{}}(entailment), \raisebox{0.2\baselineskip}{\colorbox{shallow_grey}{}}(neutral), \raisebox{0.2\baselineskip}{\colorbox{dark_yellow}{}}(contradiction).}
\label{tab_func_type}
\end{table}

Table \ref{tab_func_type} shows that the 9 com\-po\-si\-ti\-onal inference types can be grouped into 3 function types based on veridicality:\footnote{We take veridicality as example; NLI works analogously.}
i) for positive verbs $v_{e}$, the compositional inference label is consistent with the label of the NLI primitive; 
ii) for neutral verbs $v_{n}$, compositional inference remains neutral regardless of the NLI inference type;
iii) for negative verbs $v_{c}$, the compositional inference label
is the inverse of the
customary NLI label. The respective function types f$_{v_x}$ are
defined in Table \ref{tab_func_type}.
We determine the difficulty of the individual functions by averaging the results of the individual inferences pertaining to  each veridicality label $x$ in the CGen setup.  
Table \ref{tab_func_type} shows that the  performance of the 3 functions varies considerably:
f$_{v_n}$, for neutral veridicality,
exhibits significantly higher accuracy (85.74) compared to the other ones; f$_{v_e}$ for positive veridicality performs much worse (35.76) but still better than f$_{v_c}$ for negative veridicality, with 18.51 points. 
We hence 
define two \textit{\underline{c}ompositional \underline{f}unction learning \underline{o}rders (cfo)} for Task$_{CI}$: \textit{easy $\rightarrow$ hard}: f$_{v_{n}}$f$_{v_{e}}$f$_{v_{c}}$ and  \textit{hard $\rightarrow$ easy}: f$_{v_{c}}$f$_{v_{e}}$f$_{v_{n}}$.

Following $\mathcal{S}_2$ of the learning process as of \S \ref{compos_types}, we add a stage $\mathcal{S}_3$ that only presents compositional inference training data, controlled by a   
continual data stream where the functions f$_{v_{e}}$, f$_{v_{n}}$, f$_{v_{c}}$ are arranged by degree of difficulty.
$\mathcal{S}_{3,cfo}$ in Table \ref{tab_func_cl} shows the results of C$^{2}$Gen in the two opposing orders. For fair comparison, CGen is also trained with this data, 
yet in random order,
achieving 48.64 accuracy. Indeed, 
 applying the \textit{easy $\rightarrow$ hard} learning order narrows the gap to CGen up to a small margin of $\Delta$0.42, outperforming \textit{hard $\rightarrow$ easy} considerably ($\Delta$2.58). 
This finding indicates that \textbf{further training with a favorable function learning order benefits C$^{2}$Gen}, 
aligning with our insight from \S \ref{compos_types}, 
that learning easy components first enhances learning performance. 

\begin{table}[t]
\centering
\resizebox{\columnwidth}{!}{
\begin{tabular}{@{}lccc@{}} \toprule
&\multirow{2}{*}{CGen} &C$^{2}$Gen &C$^{2}$Gen  \\ 
&&easy$\rightarrow$\ hard: f$_{v_{n}}$f$_{v_{e}}$f$_{v_{c}}$ &hard$\rightarrow$\ easy:  f$_{v_{c}}$f$_{v_{e}}$f$_{v_{n}}$  \\  \midrule
$\mathcal{S}_{2,v \rightarrow n}$ &46.67 & \multicolumn{2}{c}{44.89 \footnotesize{($\Delta$ 1.78})} \\
$\mathcal{S}_{3,~cfo}$ &48.64 &48.22 (\footnotesize{$\Delta$0.42}) &46.06 (\footnotesize{$\Delta$2.58}) \\ \midrule
$\mathcal{S}_{3}^{~p ~\mid ~c_{cfo}}$ &47.19 &45.45 (\footnotesize{$\Delta$1.74}) &44.63(\footnotesize{$\Delta$2.56}) \\ \bottomrule
\end{tabular}
}
\caption{C$^2$Gen Task$_{CI}$ accuracy with \textit{\underline{c}om\-po\-si\-tio\-nal \underline{f}unction \underline{o}rdering} in two $\mathcal{S}_3$ settings (row 2-3).  
$\Delta$ shows the gap to $CGen$ for respective stages $\mathcal{S}_i$.}
\label{tab_func_cl}
\end{table}

To further consolidate the above finding we conduct a complementary experiment $\mathcal{S}_{3}^{~p ~\mid ~c_{cfo}}$. Here, we construct a 
learning scheme that follows \textit{easy before hard} but strictly orders \textit{primitive before compositional} inference. That is, the model is forced to learn independent primitive inference first, and later compositional inferences ordered by function difficulty. Row 4 in Table \ref{tab_func_cl} indicates that \textit{easy $\rightarrow$ hard} still improves over the reverse order, confirming the \textit{easy before hard} scheme. We also note that $\mathcal{S}_{3}^{~p ~\mid ~c_{cfo}}$ yields a larger gap compared to $\mathcal{S}_{3,cfo}$ (1.74 vs.\ 0.42). This suggests learning $CI$ in parallel to $P$ in $\mathcal{S}_1$, $\mathcal{S}_2$ is beneficial.

\begin{table}[t]
\centering
\resizebox{\columnwidth}{!}{
\begin{tabular}{@{}lp{7.5cm}@{}} \toprule
\makecell[l]{} & words replacement (\textcolor{card_green}{natural word}: \textcolor{dark_red}{artifical word})\\ \midrule
ver  &  \textcolor{card_green}{manage}: \textcolor{dark_red}{blicke}, \textcolor{card_green}{begin}: \textcolor{dark_red}{dmaop}, \textcolor{card_green}{hope}: \textcolor{dark_red}{lugi}, \textcolor{card_green}{wish}: \textcolor{dark_red}{fepo}, \textcolor{card_green}{expect}: \textcolor{dark_red}{kikioa}, \textcolor{card_green}{fail}: \textcolor{dark_red}{mfkd}, \textcolor{card_green}{refuse}: \textcolor{dark_red}{qneopl}
 \\ \midrule
nat & \textcolor{card_green}{coat-jacket}: \textcolor{dark_red}{nlnx-walhra}, \textcolor{card_green}{person-man}: \textcolor{dark_red}{fibqpc-qpj}, \textcolor{card_green}{rapidly-quickly}: \textcolor{dark_red}{sxaokpw-zssgjuk}, \textcolor{card_green}{dog-pet}: \textcolor{dark_red}{ozf-yqj}, \textcolor{card_green}{small-big}: \textcolor{dark_red}{noquz-srv},   \textcolor{card_green}{wet-dry}: \textcolor{dark_red}{xiw-vcs},\\ \bottomrule
\end{tabular}}
\caption{
Examples of pseudo words for \textit{ver}idical verbs and semantically related terms for \textit{nat}.}
\label{tab_pseudo_replace}
\end{table}

\section{Controlling Model Size {\scriptsize \&} Data Leakage}

PLMs \citep{devlin-etal-2019-bert, liu2019roberta} have demonstrated impressive performance on many NLP tasks through pre-training on extensive data. Recent advancements in large PLMs \citep{chowdhery2023palm, touvron2023llama} have achieved
even more substantial improvements by further scaling models and data. However, this raises
concerns 
regarding the reliability of generalization evaluations: i) \textit{re.\ data}: whether evaluation data might have been encountered during pre-training; ii) \textit{re.\ model scale}: whether a scaled PLM could show emerging
compositional generalization ability.
We address these concerns in two
experiments.

\subsection{Controlling for Data Leakage}
\begin{table*}[t]
\centering
\resizebox{\columnwidth}{!}{
\begin{tabular}{@{}llcccccccc@{}} \toprule
&\multirow{3}{*}{Settings}& \multicolumn{4}{c}{original} & \multicolumn{4}{c}{pseudo} \\ \cmidrule(r){3-6} \cmidrule(r){7-10}
& & \multicolumn{3}{c}{Task$_P$} & \multirow{2}{*}{Task$_{CI}$} & \multicolumn{3}{c}{Task$_P$} & \multirow{2}{*}{Task$_{CI}$} \\ \cmidrule(r){3-5} \cmidrule(r){7-9}
&&V&N&V+N& & V&N&V+N& \\ \midrule  
&CGen &100.00 &92.92 &92.92 &46.15 &91.76 &83.55 &81.37 &39.34\\ \midrule
\multirow{2}{*}{\rotatebox{90}{\textit{v} $\rightarrow$ \textit{n}}} &C$^{2}$Gen ($\mathcal{S}_{1}$) &100.00 &- &-  &- &90.14 &- &-  &- \\ 
&C$^{2}$Gen ($\mathcal{S}_{2}$) & 81.29($\Delta$18.71) &92.48 &78.83  &37.98($\Delta$8.17)& 76.29 ($\Delta$13.85)&81.57 &73.82 &36.62($\Delta$2.72) \\ \midrule
\multirow{2}{*}{\rotatebox{90}{\textit{n} $\rightarrow$ \textit{v}}} &C$^{2}$Gen ($\mathcal{S}_{1}$) &- &93.15 &-  &- &- &82.19 &-  &-  \\ 
&C$^{2}$Gen ($\mathcal{S}_{2}$) &99.87 &73.91($\Delta$19.24) &72.42  &34.64 ($\Delta$11.51) &89.72 &59.42 ($\Delta$22.77)  &56.72 & 33.91 ($\Delta$5.43)  \\ \bottomrule
\end{tabular}}

\caption{
Performance of Task$_{P}$ and Task$_{CI}$ on original vs.\ pseudo dataset in different training settings.}
\label{tab_pseudo_res}
\end{table*}

Following \citet{lake2023human}, we construct a \textit{pseudo-compositional} inference dataset by replacing all relevant knowledge-bearing 
natural language terms with pseudo-words. For veridical inference we replace  veridical verbs with pseudo words, e.g., \textcolor{card_green}{manage} $\rightarrow$ \textcolor{dark_red}{blicke}. Table \ref{tab_pseudo_replace} shows examples.
Irrespective of these applied changes,
we leave the signatures of the original verbs untouched.
For customary NLI, we replace pairs of semantically related words that are crucial for deciding the NLI class with a pair of pseudo words. 
E.g., in `\textit{A man catches his \textcolor{card_green}{dog} $\rightarrow$ A man catches his \textcolor{card_green}{pet}}' we 
replace
$\textcolor{card_green}{dog} \rightarrow \textcolor{dark_red}{ozf}$ and $\textcolor{card_green}{pet} \rightarrow  \textcolor{dark_red}{yqj}$.
Given the difficulty of identifying crucial semantic relations in the NLI data, we select 
438 
relation pairs covering
813 NLI instances 
(again examples in Table \ref{tab_pseudo_replace}). Like veridical inference, we preserve the original inference labels. Using these \textit{pseudo} primitive inference indicators, we 
build a \textit{pseudo} Task$_{CI}$ 
dataset following the process in \S \ref{Compositional Inference}.

With this pseudo dataset we re-evaluate the performance of RoBERTa under CGen and C$^2$Gen. The results in Table \ref{tab_pseudo_res}
align well with the trends we have seen in Table \ref{tab_q1}, for the same data in natural language. This shows that the results of our generalization experiments are not affected by data seen in pre-training. Indeed, compared to CGen, C$^2$Gen NLI shows a decline for both continual order variants of primitives \textit{ver} and \textit{nat},
in both datasets. This confirms 
that compositional generalization in NLI is more challenging in a continual learning setup. 
Comparing Task$_P$ and Task$_{CI}$ with alternating orders, 
we note that \textit{ver $\rightarrow$ nat} 
outperforms \textit{nat $\rightarrow$ ver} in both datasets. 

\begin{figure*}
    \centering
    \includegraphics[width=0.95\linewidth]{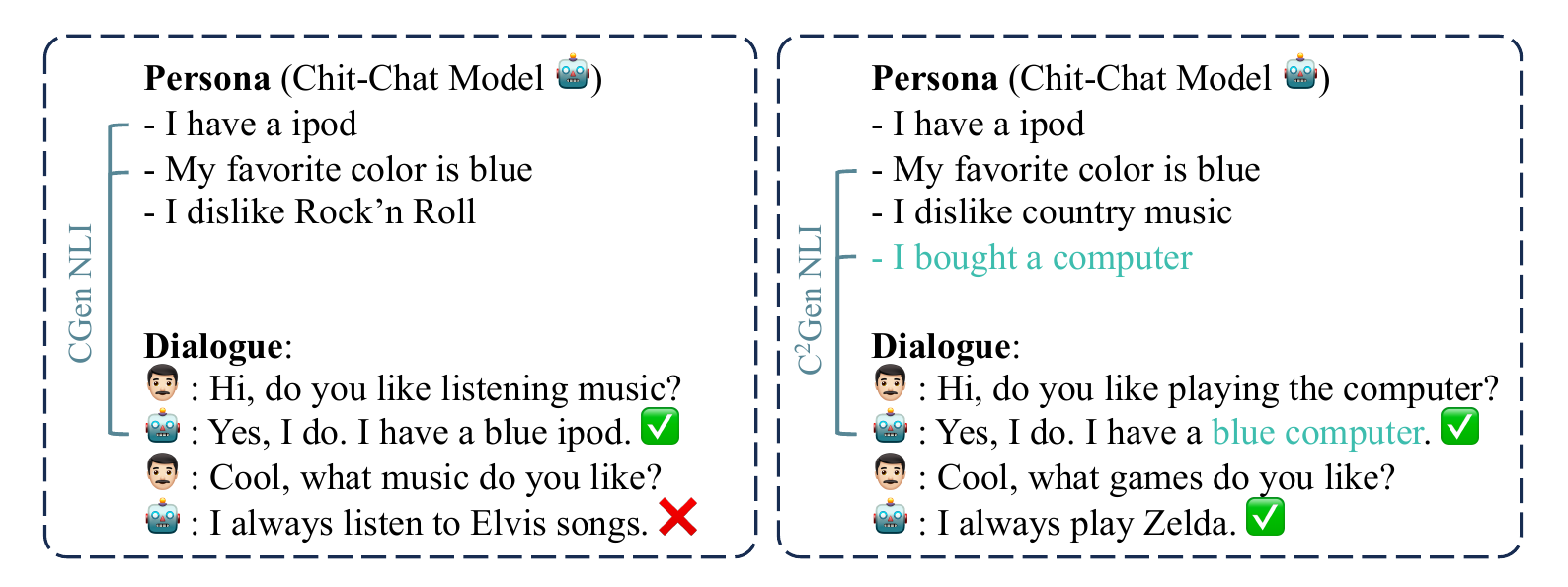}
    \caption{\textit{Persona Dialogue} application for C$^2$Gen: NLI verifies the consistency of dialogue turns generated from \textit{dynamically updated} persona information. We show a profile with \textit{new information} and compositional inferences using it (in  \raisebox{0.2\baselineskip}{\colorbox{light_green}{}}). 
} 
\label{fig:application}
\end{figure*}

Finally we observe that the absolute accuracies obtained for Task$_P$ and Task$_{CI}$ 
on \textit{pseudo} data generally drop compared to the original data, and substantially so for Task$_P$.
As for the relative performance of different continual orders regarding \textit{ver} and \textit{nat} in Task$_P$, we note that the relative drop for $nat \rightarrow ver$ compared to its opposite is much more pronounced for \textit{pseudo} vs.\ original data.

\if false
For Task$_{CI}$ this drop is milder than in Task$_{P}$, and while $ver \rightarrow nat$ still shows a small drop (\af{a.1 pp? exact?}) compared to the original data, 
$nat \rightarrow ver$ (while still somewhat lower than $ver \rightarrow nat$ in \textit{pseudo}), shows a minimal gain compared to the same setting ($nat \rightarrow ver$) on original data.

This suggests that despite masking natural with pseudo words in \textit{pseudo}, the remaining natural context can convey indicators useful for prediction, due to contextual knowledge 
stored in PLM parameters (cf.\ \citet{jawahar-etal-2019-bert, liu-etal-2021-probing-across}).

It is intuitive that leaking of implicit contextual knowledge can occur in \textit{pseudo} data, especially for customary NLI instances.
While for veridicality, single verbs are masked, for customary NLI, pseudo-words replace pairs of semantically related concepts, and therefore bear a higher risk of being revealed by surrounding context. Since Task$_{CI}$ also involves pseudo words that replace semantically related words, this holds for Task$_{CI}$ too. 

However, it is important to note that the relative performance drop compared to CGen in the original and in the pseudo setting remains unaltered. Hence, despite an asymmetry in ordering primitive tasks, the overall results of our control experiments remain unaffected: Continual learning, even if ordering primitive before compositional inference, and primitives according to relative difficulty, suffers strong losses compared to non-continual learning that cannot be attributed to models having seen test instances in pretraining.
\fi

\subsection{Model Scale: Testing C$^2$Gen  
with Llama}

\begin{table}[t]
\centering
\resizebox{\columnwidth}{!}{
\begin{tabular}{@{}llccccc@{}} \toprule
&\multirow{2}{*}{Settings} & \multicolumn{3}{c}{Task$_P$} & \multirow{2}{*}{Task$_{CI}$} \\ \cmidrule(r){3-5}
&&V&N&V+N& \\ \midrule  
&CGen &100.00 &95.17 &95.17 &49.51 \\ \midrule
\multirow{2}{*}{\rotatebox{90}{\textit{v} $\rightarrow$ \textit{n}}}&C$^{2}$Gen ($\mathcal{S}_{1}$) &100.00 &- &-  &- \\ 
&C$^{2}$Gen ($\mathcal{S}_{2}$) & 82.89($\Delta$17.11)&95.08 &79.63  & 44.39($\Delta$5.12) \\ \midrule
\multirow{2}{*}{\rotatebox{90}{\textit{n} $\rightarrow$ v}}&C$^{2}$Gen ($\mathcal{S}_{1}$) &- &95.14 &- &- \\ 
&C$^{2}$Gen ($\mathcal{S}_{2}$) &99.43 &75.24($\Delta$19.90) &74.82 &42.63($\Delta$6.88) \\ \bottomrule
\end{tabular}}
\caption{Results for Task$_P$
and Task$_{CI}$ in different training settings with Llama2-7b. }
\label{tab_llama}
\end{table}

We next test the generalization ability for C$^2$Gen NLI for a large PLM such as   
Llama-2-7B \citep{touvron2023llama}. To fine-tune this large PLM, we adopt standard parameter-efficient fine-tuning (peft) with
LoRA \citep{hu2022lora}. Compared to RoBERTa-Large, 
its size increases by approx.\ 20 times from  0.355 to 7 billion parameters. 
This enhances the 
accuracy on the CGen
test
from 46.67 to 49.51\% ($\Delta$ 2.84) for Task$_{CI}$. While this marks a progress, a large drop occurs for continual learning in C$^2$Gen ($\Delta$ 5.12/$\Delta$ 6.88).
This suggests that compositional generalization is still a challenge for LLMs. Besides, the gain of 2.84 over RoBERTa on CGen is constrained, 
compared to the substantial resource cost. This finding is consistent with \citet{qiu-etal-2022-evaluating}, who found that ﬁne-tuning LLMs generally has a ﬂat or negative scaling curve on compositional generalization in semantic parsing.

Similar to RoBERTa, we observe that Llama-2 is affected by forgetting --
but the amount of forgetting in Llama-2 
does not differ much, dropping by 1.6 points in \textit{ver $\rightarrow$ nat} but rising by 0.66 points in \textit{nat $\rightarrow$ ver}.
Comparing different training orders (\textit{ver $\rightarrow$ nat}, \textit{nat $\rightarrow$ ver}) confirms that Llama-2 also benefits from an `easy to hard' learning scheme.

\section{Potential Applications}
\label{sec:application}

Our work introduces the
new C$^{2}$Gen NLI task 
as a first step to explore the compositional generalization ability of models performing NLI in a continual learning setup. Similar to existing continual NLP-based tasks \citep{wang-etal-2019-sentence, berard-2021-continual, madotto-etal-2021-continual, mhamdi-etal-2023-cross}, the continual learning setup inspires models to learn new inference knowledge continuously, to avoid costs for
model retraining.
Given such capabilities, the
 C$^{2}$Gen NLI task setting can benefit 
 future applications that require the understanding and induction of compositional 
inferences relative to dynamically updated knowledge stores.

We use the widely researched task \textit{Personalized Dialogue Agent (PDA)} \citep{zhang-etal-2018-personalizing} as an example to show how the C$^{2}$Gen NLI task could apply in a dynamic setting.
\textit{PDA} 
proposes chit-chat models that are conditioned on information provided in a given personality profile.  Fig.\ \ref{fig:application} shows an illustration.
Existing approaches suffer from consistency issues when a chit-chat model generates utterances that contradict 
their personality profile. E.g., \textit{I dislike Rock'n Roll} contradicts
\textit{I always listen to Elvis songs}.
To solve this issue, some works \citep{welleck-etal-2019-dialogue, utama-etal-2022-falsesum} proposed to use NLI to evaluate and improve consistency. We can achieve this
by evaluating whether the persona information entails or contradicts a dialogue utterance. In dialogue, utterances 
show semantic composition effects when combining  primitive information to form new and meaningful sentences.
E.g., \textit{I have a blue iPod} 
composes information from
\textit{I have an iPod} and \textit{my favorite color is blue}. This scenario aligns with the CGen NLI setup.

But the persona profile of a chit-chat is dy\-na\-mic and 
gets 
updated over time.
For example, Fig.\ \ref{fig:application} shows 
persona information that is updated with a fact on a 
new product \textit{computer}. The new primitive can be composed 
with previously learned primitives to generate novel compositional facts, e.g., 
\textit{I have a blue computer} from
\textit{I bought a computer} and \textit{my favorite color is blue}. Here, re-training the model 
to update the profile's information state
is expensive and time-consuming. By contrast, enabling the model to perform continual learning is a more viable and economic solution. The 
model is then deemed to evaluate compositional inferences \textit{relative to the updated information state},
aligning with our new task C$^{2}$Gen NLI.

\section{Conclusions and Future Work}

We propose C$^{2}$Gen, a new challenge task for compositional generalization 
in NLI, grounded in a continual learning scenario. Our new task 
targets NLP applications that rely on composing information from continuously updated sources.

By conducting rich analyses for this novel task, on our new benchmark, we show that in continual learning, neural models fail to generalize to unseen compositional inferences due to \textit{forgetting}. With known continual learning strategies we can combat forgetting, but our analyses show that \textit{memorization alone cannot solve the compositional inference challenge}. 
Our in-depth analyses of  C$^{2}$Gen show that the model benefits from \textit{learning primitive before compositional inference}, and \textit{learning easy before hard inference subtasks}.

Our findings highlight the importance of 
observing differences of primitive and compositional inference types, 
and establishing the relative difficulties of
diverse primitive and compositional inference types. 
With this, we establish recipes that can improve continual learning 
to approach non-continual learning. New methods can 
determine optimal learning orders for diverse inference types, while ensuring sufficient 
diversity in the data stream.
Our insights could also benefit other compositional generalization methods, e.g., by ordering demonstrations in in-context learning along principles we established to improve compositional generalization in continual learning.

\subsection*{Acknowledgements} We are grateful to the anonymous reviewers, and Action Editors Mihai Surdeanu and Katrin Elisabeth Erk for their valuable comments. This work has been supported through a scholarship provided by the Heidelberg Institute for Theoretical Studies gGmbH.

\bibliography{tacl2021,anthology}

\begin{thebibliography}{72}
\expandafter\ifx\csname natexlab\endcsname\relax\def\natexlab#1{#1}\fi

\bibitem[{Aguilar et~al.(2020)Aguilar, Ling, Zhang, Yao, Fan, and Guo}]{aguilar2020knowledge}
Gustavo Aguilar, Yuan Ling, Yu~Zhang, Benjamin Yao, Xing Fan, and Chenlei Guo. 2020.
\newblock Knowledge distillation from internal representations.
\newblock In \emph{Proceedings of the AAAI Conference on Artificial Intelligence}, volume~34, pages 7350--7357.

\bibitem[{Aljundi et~al.(2019)Aljundi, Belilovsky, Tuytelaars, Charlin, Caccia, Lin, and Page-Caccia}]{Aljundi2019online}
Rahaf Aljundi, Eugene Belilovsky, Tinne Tuytelaars, Laurent Charlin, Massimo Caccia, Min Lin, and Lucas Page-Caccia. 2019.
\newblock Online continual learning with maximal interfered retrieval.
\newblock In \emph{Advances in Neural Information Processing Systems 32}.

\bibitem[{Bauer et~al.(2021)Bauer, Deng, and Bansal}]{bauer-etal-2021-ernie}
Lisa Bauer, Lingjia Deng, and Mohit Bansal. 2021.
\newblock \href {https://doi.org/10.18653/v1/2021.deelio-1.7} {{ERNIE}-{NLI}: Analyzing the impact of domain-specific external knowledge on enhanced representations for {NLI}}.
\newblock In \emph{Proceedings of Deep Learning Inside Out (DeeLIO): The 2nd Workshop on Knowledge Extraction and Integration for Deep Learning Architectures}, pages 58--69, Online. Association for Computational Linguistics.

\bibitem[{Bengio et~al.(2009)Bengio, Louradour, Collobert, and Weston}]{bengio2009curriculum}
Yoshua Bengio, J{\'e}r{\^o}me Louradour, Ronan Collobert, and Jason Weston. 2009.
\newblock Curriculum learning.
\newblock In \emph{Proceedings of the 26th annual international conference on machine learning}, pages 41--48.

\bibitem[{Berard(2021)}]{berard-2021-continual}
Alexandre Berard. 2021.
\newblock \href {https://aclanthology.org/2021.wmt-1.62} {Continual learning in multilingual {NMT} via language-specific embeddings}.
\newblock In \emph{Proceedings of the Sixth Conference on Machine Translation}, pages 542--565, Online. Association for Computational Linguistics.

\bibitem[{Berglund et~al.(2024)Berglund, Tong, Kaufmann, Balesni, Stickland, Korbak, and Evans}]{berglund2023reversal}
Lukas Berglund, Meg Tong, Maximilian Kaufmann, Mikita Balesni, Asa~Cooper Stickland, Tomasz Korbak, and Owain Evans. 2024.
\newblock The reversal curse: {LLM}s trained on {\textquotedblleft}a is b{\textquotedblright} fail to learn {\textquotedblleft}b is a{\textquotedblright}.
\newblock In \emph{International Conference on Learning Representations}.

\bibitem[{Bhargava et~al.(2021)Bhargava, Drozd, and Rogers}]{bhargava-etal-2021-generalization}
Prajjwal Bhargava, Aleksandr Drozd, and Anna Rogers. 2021.
\newblock \href {https://doi.org/10.18653/v1/2021.insights-1.18} {Generalization in {NLI}: Ways (not) to go beyond simple heuristics}.
\newblock In \emph{Proceedings of the Second Workshop on Insights from Negative Results in NLP}, pages 125--135, Online and Punta Cana, Dominican Republic. Association for Computational Linguistics.

\bibitem[{Biesialska et~al.(2020)Biesialska, Biesialska, and Costa-juss{\`a}}]{biesialska-etal-2020-continual}
Magdalena Biesialska, Katarzyna Biesialska, and Marta~R. Costa-juss{\`a}. 2020.
\newblock \href {https://doi.org/10.18653/v1/2020.coling-main.574} {Continual lifelong learning in natural language processing: A survey}.
\newblock In \emph{Proceedings of the 28th International Conference on Computational Linguistics}, pages 6523--6541, Barcelona, Spain (Online). International Committee on Computational Linguistics.

\bibitem[{Bowman et~al.(2015)Bowman, Angeli, Potts, and Manning}]{bowman-etal-2015-large}
Samuel~R. Bowman, Gabor Angeli, Christopher Potts, and Christopher~D. Manning. 2015.
\newblock \href {https://doi.org/10.18653/v1/D15-1075} {A large annotated corpus for learning natural language inference}.
\newblock In \emph{Proceedings of the 2015 Conference on Empirical Methods in Natural Language Processing}, pages 632--642, Lisbon, Portugal. Association for Computational Linguistics.

\bibitem[{Chaudhry et~al.(2019{\natexlab{a}})Chaudhry, Ranzato, Rohrbach, and Elhoseiny}]{chaudhry2019efficient}
Arslan Chaudhry, Marc'Aurelio Ranzato, Marcus Rohrbach, and Mohamed Elhoseiny. 2019{\natexlab{a}}.
\newblock Efficient lifelong learning with a-gem.
\newblock In \emph{International Conference on Learning Representations}.

\bibitem[{Chaudhry et~al.(2019{\natexlab{b}})Chaudhry, Rohrbach, Elhoseiny, Ajanthan, Dokania, Torr, and Ranzato}]{Chaudhry2019OnTE}
Arslan Chaudhry, Marcus Rohrbach, Mohamed Elhoseiny, Thalaiyasingam Ajanthan, Puneet~Kumar Dokania, Philip H.~S. Torr, and Marc'Aurelio Ranzato. 2019{\natexlab{b}}.
\newblock On tiny episodic memories in continual learning.
\newblock \emph{arXiv: Learning}.

\bibitem[{Chen et~al.(2021)Chen, Choi, and Durrett}]{chen-etal-2021-nli-models}
Jifan Chen, Eunsol Choi, and Greg Durrett. 2021.
\newblock \href {https://doi.org/10.18653/v1/2021.findings-emnlp.324} {Can {NLI} models verify {QA} systems{'} predictions?}
\newblock In \emph{Findings of the Association for Computational Linguistics: EMNLP 2021}, pages 3841--3854, Punta Cana, Dominican Republic. Association for Computational Linguistics.

\bibitem[{Chen et~al.(2018)Chen, Zhu, Ling, Inkpen, and Wei}]{chen-etal-2018-neural-natural}
Qian Chen, Xiaodan Zhu, Zhen-Hua Ling, Diana Inkpen, and Si~Wei. 2018.
\newblock \href {https://doi.org/10.18653/v1/P18-1224} {Neural natural language inference models enhanced with external knowledge}.
\newblock In \emph{Proceedings of the 56th Annual Meeting of the Association for Computational Linguistics (Volume 1: Long Papers)}, pages 2406--2417, Melbourne, Australia. Association for Computational Linguistics.

\bibitem[{Chowdhery et~al.(2023)Chowdhery, Narang, Devlin, Bosma, Mishra, Roberts, Barham, Chung, Sutton, Gehrmann et~al.}]{chowdhery2023palm}
Aakanksha Chowdhery, Sharan Narang, Jacob Devlin, Maarten Bosma, Gaurav Mishra, Adam Roberts, Paul Barham, Hyung~Won Chung, Charles Sutton, Sebastian Gehrmann, et~al. 2023.
\newblock Palm: Scaling language modeling with pathways.
\newblock \emph{Journal of Machine Learning Research}, 24(240):1--113.

\bibitem[{Conneau et~al.(2018)Conneau, Rinott, Lample, Williams, Bowman, Schwenk, and Stoyanov}]{conneau-etal-2018-xnli}
Alexis Conneau, Ruty Rinott, Guillaume Lample, Adina Williams, Samuel Bowman, Holger Schwenk, and Veselin Stoyanov. 2018.
\newblock \href {https://doi.org/10.18653/v1/D18-1269} {{XNLI}: Evaluating cross-lingual sentence representations}.
\newblock In \emph{Proceedings of the 2018 Conference on Empirical Methods in Natural Language Processing}, pages 2475--2485, Brussels, Belgium. Association for Computational Linguistics.

\bibitem[{Dagan et~al.(2013)Dagan, Roth, Sammons, and Zanzotto}]{dagan2013recognizing}
Ido Dagan, Dan Roth, Mark Sammons, and Fabio~Massimo Zanzotto. 2013.
\newblock Recognizing textual entailment: Models and applications.
\newblock \emph{Synthesis Lectures on Human Language Technologies}, 6(4):1--220.

\bibitem[{Devlin et~al.(2019)Devlin, Chang, Lee, and Toutanova}]{devlin-etal-2019-bert}
Jacob Devlin, Ming-Wei Chang, Kenton Lee, and Kristina Toutanova. 2019.
\newblock \href {https://doi.org/10.18653/v1/N19-1423} {{BERT}: Pre-training of deep bidirectional transformers for language understanding}.
\newblock In \emph{Proceedings of the 2019 Conference of the North {A}merican Chapter of the Association for Computational Linguistics: Human Language Technologies, Volume 1 (Long and Short Papers)}, pages 4171--4186, Minneapolis, Minnesota. Association for Computational Linguistics.

\bibitem[{Dziri et~al.(2023)Dziri, Lu, Sclar, Li, Jiang, Lin, Welleck, West, Bhagavatula, Bras, Hwang, Sanyal, Ren, Ettinger, Harchaoui, and Choi}]{dziri2023faith}
Nouha Dziri, Ximing Lu, Melanie Sclar, Xiang~Lorraine Li, Liwei Jiang, Bill~Yuchen Lin, Sean Welleck, Peter West, Chandra Bhagavatula, Ronan~Le Bras, Jena~D. Hwang, Soumya Sanyal, Xiang Ren, Allyson Ettinger, Zaid Harchaoui, and Yejin Choi. 2023.
\newblock \href {https://openreview.net/forum?id=Fkckkr3ya8} {Faith and fate: Limits of transformers on compositionality}.
\newblock In \emph{Thirty-seventh Conference on Neural Information Processing Systems}.

\bibitem[{Elman(1993)}]{elman1993learning}
Jeffrey~L Elman. 1993.
\newblock Learning and development in neural networks: The importance of starting small.
\newblock \emph{Cognition}, 48(1):71--99.

\bibitem[{Fodor and Pylyshyn(1988)}]{fodor1988connectionism}
Jerry~A Fodor and Zenon~W Pylyshyn. 1988.
\newblock Connectionism and cognitive architecture: A critical analysis.
\newblock \emph{Cognition}, 28(1-2):3--71.

\bibitem[{French(1999)}]{french1999catastrophic}
Robert~M French. 1999.
\newblock Catastrophic forgetting in connectionist networks.
\newblock \emph{Trends in cognitive sciences}, 3(4):128--135.

\bibitem[{Fu and Frank(2023)}]{fu-frank-2023-seti}
Xiyan Fu and Anette Frank. 2023.
\newblock \href {https://doi.org/10.18653/v1/2023.findings-acl.252} {{SETI}: Systematicity evaluation of textual inference}.
\newblock In \emph{Findings of the Association for Computational Linguistics: ACL 2023}, pages 4101--4114, Toronto, Canada. Association for Computational Linguistics.

\bibitem[{Geiger et~al.(2020)Geiger, Richardson, and Potts}]{geiger-etal-2020-neural}
Atticus Geiger, Kyle Richardson, and Christopher Potts. 2020.
\newblock \href {https://doi.org/10.18653/v1/2020.blackboxnlp-1.16} {Neural natural language inference models partially embed theories of lexical entailment and negation}.
\newblock In \emph{Proceedings of the Third BlackboxNLP Workshop on Analyzing and Interpreting Neural Networks for NLP}, pages 163--173, Online. Association for Computational Linguistics.

\bibitem[{Gong et~al.(2018)Gong, Luo, and Zhang}]{gong2017natural}
Yichen Gong, Heng Luo, and Jian Zhang. 2018.
\newblock Natural language inference over interaction space.
\newblock In \emph{International Conference on Learning Representations}.

\bibitem[{Goodwin et~al.(2020)Goodwin, Sinha, and O{'}Donnell}]{goodwin-etal-2020-probing}
Emily Goodwin, Koustuv Sinha, and Timothy~J. O{'}Donnell. 2020.
\newblock \href {https://doi.org/10.18653/v1/2020.acl-main.177} {Probing linguistic systematicity}.
\newblock In \emph{Proceedings of the 58th Annual Meeting of the Association for Computational Linguistics}, pages 1958--1969, Online. Association for Computational Linguistics.

\bibitem[{Hu et~al.(2022)Hu, Shen, Wallis, Allen-Zhu, Li, Wang, Wang, and Chen}]{hu2022lora}
Edward~J Hu, Yelong Shen, Phillip Wallis, Zeyuan Allen-Zhu, Yuanzhi Li, Shean Wang, Lu~Wang, and Weizhu Chen. 2022.
\newblock \href {https://openreview.net/forum?id=nZeVKeeFYf9} {Lo{RA}: Low-rank adaptation of large language models}.
\newblock In \emph{International Conference on Learning Representations}.

\bibitem[{Huang and Du(2019)}]{huang-du-2019-self}
Yuyun Huang and Jinhua Du. 2019.
\newblock \href {https://doi.org/10.18653/v1/D19-1037} {Self-attention enhanced {CNN}s and collaborative curriculum learning for distantly supervised relation extraction}.
\newblock In \emph{Proceedings of the 2019 Conference on Empirical Methods in Natural Language Processing and the 9th International Joint Conference on Natural Language Processing (EMNLP-IJCNLP)}, pages 389--398, Hong Kong, China. Association for Computational Linguistics.

\bibitem[{Hupkes et~al.(2020)Hupkes, Dankers, Mul, and Bruni}]{hupkes2020compositionality}
Dieuwke Hupkes, Verna Dankers, Mathijs Mul, and Elia Bruni. 2020.
\newblock Compositionality decomposed: How do neural networks generalise?
\newblock \emph{Journal of Artificial Intelligence Research}, 67:757--795.

\bibitem[{Jia et~al.(2023)Jia, Liu, Tang, and Zhu}]{jia-etal-2023-sample}
Qi~Jia, Yizhu Liu, Haifeng Tang, and Kenny Zhu. 2023.
\newblock \href {https://doi.org/10.18653/v1/2023.acl-long.666} {In-sample curriculum learning by sequence completion for natural language generation}.
\newblock In \emph{Proceedings of the 61st Annual Meeting of the Association for Computational Linguistics (Volume 1: Long Papers)}, pages 11937--11950, Toronto, Canada. Association for Computational Linguistics.

\bibitem[{Jiang and de~Marneffe(2022)}]{jiang-marneffe-2022-investigating}
Nan-Jiang Jiang and Marie-Catherine de~Marneffe. 2022.
\newblock \href {https://doi.org/10.1162/tacl_a_00523} {Investigating reasons for disagreement in natural language inference}.
\newblock \emph{Transactions of the Association for Computational Linguistics}, 10:1357--1374.

\bibitem[{Jin et~al.(2020)Jin, Du, Sadhu, Nevatia, and Ren}]{jin-etal-2020-visually}
Xisen Jin, Junyi Du, Arka Sadhu, Ram Nevatia, and Xiang Ren. 2020.
\newblock \href {https://doi.org/10.18653/v1/2020.emnlp-main.158} {Visually grounded continual learning of compositional phrases}.
\newblock In \emph{Proceedings of the 2020 Conference on Empirical Methods in Natural Language Processing (EMNLP)}, pages 2018--2029, Online. Association for Computational Linguistics.

\bibitem[{Kalouli et~al.(2023)Kalouli, Hu, Webb, Moss, and De~Paiva}]{kalouli2023curing}
Aikaterini-Lida Kalouli, Hai Hu, Alexander~F Webb, Lawrence~S Moss, and Valeria De~Paiva. 2023.
\newblock Curing the sick and other nli maladies.
\newblock \emph{Computational Linguistics}, 49(1):199--243.

\bibitem[{Karttunen(1971)}]{karttunen1971implicative}
Lauri Karttunen. 1971.
\newblock Implicative verbs.
\newblock \emph{Language}, pages 340--358.

\bibitem[{Keysers et~al.(2020)Keysers, Sch{\"a}rli, Scales, Buisman, Furrer, Kashubin, Momchev, Sinopalnikov, Stafiniak, Tihon et~al.}]{keysers2020measuring}
Daniel Keysers, Nathanael Sch{\"a}rli, Nathan Scales, Hylke Buisman, Daniel Furrer, Sergii Kashubin, Nikola Momchev, Danila Sinopalnikov, Lukasz Stafiniak, Tibor Tihon, et~al. 2020.
\newblock Measuring compositional generalization: A comprehensive method on realistic data.
\newblock In \emph{International Conference on Learning Representations}.

\bibitem[{Krueger and Dayan(2009)}]{krueger2009flexible}
Kai~A Krueger and Peter Dayan. 2009.
\newblock Flexible shaping: How learning in small steps helps.
\newblock \emph{Cognition}, 110(3):380--394.

\bibitem[{Kumar et~al.(2010)Kumar, Packer, and Koller}]{kumar2010self}
M~Kumar, Benjamin Packer, and Daphne Koller. 2010.
\newblock Self-paced learning for latent variable models.
\newblock \emph{Advances in neural information processing systems}, 23.

\bibitem[{Laban et~al.(2022)Laban, Schnabel, Bennett, and Hearst}]{laban-etal-2022-summac}
Philippe Laban, Tobias Schnabel, Paul~N. Bennett, and Marti~A. Hearst. 2022.
\newblock \href {https://doi.org/10.1162/tacl_a_00453} {{S}umma{C}: Re-visiting {NLI}-based models for inconsistency detection in summarization}.
\newblock \emph{Transactions of the Association for Computational Linguistics}, 10:163--177.

\bibitem[{Lai et~al.(2017)Lai, Bisk, and Hockenmaier}]{lai-etal-2017-natural}
Alice Lai, Yonatan Bisk, and Julia Hockenmaier. 2017.
\newblock \href {https://aclanthology.org/I17-1011} {Natural language inference from multiple premises}.
\newblock In \emph{Proceedings of the Eighth International Joint Conference on Natural Language Processing (Volume 1: Long Papers)}, pages 100--109, Taipei, Taiwan. Asian Federation of Natural Language Processing.

\bibitem[{Lake and Baroni(2023)}]{lake2023human}
Brenden~M Lake and Marco Baroni. 2023.
\newblock Human-like systematic generalization through a meta-learning neural network.
\newblock \emph{Nature}, pages 1--7.

\bibitem[{Li et~al.(2020)Li, Zhao, Church, and Elhoseiny}]{li2020compositional}
Yuanpeng Li, Liang Zhao, Kenneth Church, and Mohamed Elhoseiny. 2020.
\newblock Compositional language continual learning.
\newblock In \emph{International Conference on Learning Representations}.

\bibitem[{Liu et~al.(2022)Liu, Lewis, Riedel, and Stenetorp}]{liu-etal-2022-challenges}
Linqing Liu, Patrick Lewis, Sebastian Riedel, and Pontus Stenetorp. 2022.
\newblock \href {https://doi.org/10.18653/v1/2022.findings-naacl.155} {Challenges in generalization in open domain question answering}.
\newblock In \emph{Findings of the Association for Computational Linguistics: NAACL 2022}, pages 2014--2029, Seattle, United States. Association for Computational Linguistics.

\bibitem[{Liu et~al.(2024)Liu, Wan, and Strube}]{liu2024causes}
Wei Liu, Stephen Wan, and Michael Strube. 2024.
\newblock What causes the failure of explicit to implicit discourse relation recognition?
\newblock \emph{arXiv preprint arXiv:2404.00999}.

\bibitem[{Liu et~al.(2019)Liu, Ott, Goyal, Du, Joshi, Chen, Levy, Lewis, Zettlemoyer, and Stoyanov}]{liu2019roberta}
Yinhan Liu, Myle Ott, Naman Goyal, Jingfei Du, Mandar Joshi, Danqi Chen, Omer Levy, Mike Lewis, Luke Zettlemoyer, and Veselin Stoyanov. 2019.
\newblock Roberta: A robustly optimized bert pretraining approach.
\newblock \emph{arXiv preprint arXiv:1907.11692}.

\bibitem[{Madotto et~al.(2021)Madotto, Lin, Zhou, Moon, Crook, Liu, Yu, Cho, Fung, and Wang}]{madotto-etal-2021-continual}
Andrea Madotto, Zhaojiang Lin, Zhenpeng Zhou, Seungwhan Moon, Paul Crook, Bing Liu, Zhou Yu, Eunjoon Cho, Pascale Fung, and Zhiguang Wang. 2021.
\newblock \href {https://doi.org/10.18653/v1/2021.emnlp-main.590} {Continual learning in task-oriented dialogue systems}.
\newblock In \emph{Proceedings of the 2021 Conference on Empirical Methods in Natural Language Processing}, pages 7452--7467, Online and Punta Cana, Dominican Republic. Association for Computational Linguistics.

\bibitem[{Marelli et~al.(2014)Marelli, Menini, Baroni, Bentivogli, Bernardi, and Zamparelli}]{marelli-etal-2014-sick}
Marco Marelli, Stefano Menini, Marco Baroni, Luisa Bentivogli, Raffaella Bernardi, and Roberto Zamparelli. 2014.
\newblock \href {http://www.lrec-conf.org/proceedings/lrec2014/pdf/363_Paper.pdf} {A {SICK} cure for the evaluation of compositional distributional semantic models}.
\newblock In \emph{Proceedings of the Ninth International Conference on Language Resources and Evaluation ({LREC}'14)}, pages 216--223, Reykjavik, Iceland. European Language Resources Association (ELRA).

\bibitem[{McCloskey and Cohen(1989)}]{mccloskey1989connectionist}
Michael McCloskey and Neal~J Cohen. 1989.
\newblock Catastrophic interference in connectionist networks: The sequential learning problem.
\newblock In \emph{Psychology of learning and motivation}, volume~24, pages 109--165. Elsevier.

\bibitem[{McCoy et~al.(2019)McCoy, Pavlick, and Linzen}]{mccoy-etal-2019-right}
Tom McCoy, Ellie Pavlick, and Tal Linzen. 2019.
\newblock \href {https://doi.org/10.18653/v1/P19-1334} {Right for the wrong reasons: Diagnosing syntactic heuristics in natural language inference}.
\newblock In \emph{Proceedings of the 57th Annual Meeting of the Association for Computational Linguistics}, pages 3428--3448, Florence, Italy. Association for Computational Linguistics.

\bibitem[{M{'}hamdi et~al.(2023)M{'}hamdi, Ren, and May}]{mhamdi-etal-2023-cross}
Meryem M{'}hamdi, Xiang Ren, and Jonathan May. 2023.
\newblock \href {https://doi.org/10.18653/v1/2023.acl-long.217} {Cross-lingual continual learning}.
\newblock In \emph{Proceedings of the 61st Annual Meeting of the Association for Computational Linguistics (Volume 1: Long Papers)}, pages 3908--3943, Toronto, Canada. Association for Computational Linguistics.

\bibitem[{Mitchell et~al.(2023)Mitchell, Palmarini, and Moskvichev}]{mitchell2023comparing}
Melanie Mitchell, Alessandro~B. Palmarini, and Arsenii~Kirillovich Moskvichev. 2023.
\newblock Comparing humans, {GPT}-4, and {GPT}-4v on abstraction and reasoning tasks.
\newblock In \emph{AAAI 2024 Workshop on ''Are Large Language Models Simply Causal Parrots?''}.

\bibitem[{Nie et~al.(2020)Nie, Zhou, and Bansal}]{nie-etal-2020-learn}
Yixin Nie, Xiang Zhou, and Mohit Bansal. 2020.
\newblock \href {https://doi.org/10.18653/v1/2020.emnlp-main.734} {What can we learn from collective human opinions on natural language inference data?}
\newblock In \emph{Proceedings of the 2020 Conference on Empirical Methods in Natural Language Processing (EMNLP)}, pages 9131--9143, Online. Association for Computational Linguistics.

\bibitem[{Parikh et~al.(2016)Parikh, T{\"a}ckstr{\"o}m, Das, and Uszkoreit}]{parikh-etal-2016-decomposable}
Ankur Parikh, Oscar T{\"a}ckstr{\"o}m, Dipanjan Das, and Jakob Uszkoreit. 2016.
\newblock \href {https://doi.org/10.18653/v1/D16-1244} {A decomposable attention model for natural language inference}.
\newblock In \emph{Proceedings of the 2016 Conference on Empirical Methods in Natural Language Processing}, pages 2249--2255, Austin, Texas. Association for Computational Linguistics.

\bibitem[{Parisi et~al.(2019)Parisi, Kemker, Part, Kanan, and Wermter}]{parisi2019continual}
German~I Parisi, Ronald Kemker, Jose~L Part, Christopher Kanan, and Stefan Wermter. 2019.
\newblock Continual lifelong learning with neural networks: A review.
\newblock \emph{Neural networks}, 113:54--71.

\bibitem[{Pavlick and Kwiatkowski(2019)}]{pavlick-kwiatkowski-2019-inherent}
Ellie Pavlick and Tom Kwiatkowski. 2019.
\newblock \href {https://doi.org/10.1162/tacl_a_00293} {Inherent disagreements in human textual inferences}.
\newblock \emph{Transactions of the Association for Computational Linguistics}, 7:677--694.

\bibitem[{Qiu et~al.(2022)Qiu, Shaw, Pasupat, Shi, Herzig, Pitler, Sha, and Toutanova}]{qiu-etal-2022-evaluating}
Linlu Qiu, Peter Shaw, Panupong Pasupat, Tianze Shi, Jonathan Herzig, Emily Pitler, Fei Sha, and Kristina Toutanova. 2022.
\newblock \href {https://aclanthology.org/2022.emnlp-main.624} {Evaluating the impact of model scale for compositional generalization in semantic parsing}.
\newblock In \emph{Proceedings of the 2022 Conference on Empirical Methods in Natural Language Processing}, pages 9157--9179, Abu Dhabi, United Arab Emirates. Association for Computational Linguistics.

\bibitem[{Raffel et~al.(2020)Raffel, Shazeer, Roberts, Lee, Narang, Matena, Zhou, Li, and Liu}]{raffel2020exploring}
Colin Raffel, Noam Shazeer, Adam Roberts, Katherine Lee, Sharan Narang, Michael Matena, Yanqi Zhou, Wei Li, and Peter~J Liu. 2020.
\newblock Exploring the limits of transfer learning with a unified text-to-text transformer.
\newblock \emph{The Journal of Machine Learning Research}, 21(1):5485--5551.

\bibitem[{Ranaldi et~al.(2023)Ranaldi, Pucci, and Zanzotto}]{ranaldi-etal-2023-modeling}
Leonardo Ranaldi, Giulia Pucci, and Fabio~Massimo Zanzotto. 2023.
\newblock \href {https://aclanthology.org/2023.ranlp-1.101} {Modeling easiness for training transformers with curriculum learning}.
\newblock In \emph{Proceedings of the 14th International Conference on Recent Advances in Natural Language Processing}, pages 937--948, Varna, Bulgaria. INCOMA Ltd., Shoumen, Bulgaria.

\bibitem[{Ring(1997)}]{ring1997child}
Mark~B Ring. 1997.
\newblock Child: A first step towards continual learning.
\newblock \emph{Machine Learning}, 28:77--104.

\bibitem[{Robins(1995)}]{robins1995catastrophic}
Anthony Robins. 1995.
\newblock Catastrophic forgetting, rehearsal and pseudorehearsal.
\newblock \emph{Connection Science}, 7(2):123--146.

\bibitem[{Ross and Pavlick(2019)}]{ross-pavlick-2019-well}
Alexis Ross and Ellie Pavlick. 2019.
\newblock \href {https://doi.org/10.18653/v1/D19-1228} {How well do {NLI} models capture verb veridicality?}
\newblock In \emph{Proceedings of the 2019 Conference on Empirical Methods in Natural Language Processing and the 9th International Joint Conference on Natural Language Processing (EMNLP-IJCNLP)}, pages 2230--2240, Hong Kong, China. Association for Computational Linguistics.

\bibitem[{Soviany et~al.(2022)Soviany, Ionescu, Rota, and Sebe}]{soviany2022curriculum}
Petru Soviany, Radu~Tudor Ionescu, Paolo Rota, and Nicu Sebe. 2022.
\newblock Curriculum learning: A survey.
\newblock \emph{International Journal of Computer Vision}, 130(6):1526--1565.

\bibitem[{Stasaski and Hearst(2022)}]{stasaski-hearst-2022-semantic}
Katherine Stasaski and Marti Hearst. 2022.
\newblock \href {https://doi.org/10.18653/v1/2022.naacl-main.6} {Semantic diversity in dialogue with natural language inference}.
\newblock In \emph{Proceedings of the 2022 Conference of the North American Chapter of the Association for Computational Linguistics: Human Language Technologies}, pages 85--98, Seattle, United States. Association for Computational Linguistics.

\bibitem[{Touvron et~al.(2023)Touvron, Martin, Stone, Albert, Almahairi, Babaei, Bashlykov, Batra, Bhargava, Bhosale et~al.}]{touvron2023llama}
Hugo Touvron, Louis Martin, Kevin Stone, Peter Albert, Amjad Almahairi, Yasmine Babaei, Nikolay Bashlykov, Soumya Batra, Prajjwal Bhargava, Shruti Bhosale, et~al. 2023.
\newblock Llama 2: Open foundation and fine-tuned chat models.
\newblock \emph{arXiv preprint arXiv:2307.09288}.

\bibitem[{Utama et~al.(2022)Utama, Bambrick, Moosavi, and Gurevych}]{utama-etal-2022-falsesum}
Prasetya Utama, Joshua Bambrick, Nafise Moosavi, and Iryna Gurevych. 2022.
\newblock \href {https://doi.org/10.18653/v1/2022.naacl-main.199} {Falsesum: Generating document-level {NLI} examples for recognizing factual inconsistency in summarization}.
\newblock In \emph{Proceedings of the 2022 Conference of the North American Chapter of the Association for Computational Linguistics: Human Language Technologies}, pages 2763--2776, Seattle, United States. Association for Computational Linguistics.

\bibitem[{Wang et~al.(2019{\natexlab{a}})Wang, Pruksachatkun, Nangia, Singh, Michael, Hill, Levy, and Bowman}]{wang2019superglue}
Alex Wang, Yada Pruksachatkun, Nikita Nangia, Amanpreet Singh, Julian Michael, Felix Hill, Omer Levy, and Samuel Bowman. 2019{\natexlab{a}}.
\newblock Superglue: A stickier benchmark for general-purpose language understanding systems.
\newblock \emph{Advances in neural information processing systems}, 32.

\bibitem[{Wang et~al.(2019{\natexlab{b}})Wang, Xiong, Yu, Guo, Chang, and Wang}]{wang-etal-2019-sentence}
Hong Wang, Wenhan Xiong, Mo~Yu, Xiaoxiao Guo, Shiyu Chang, and William~Yang Wang. 2019{\natexlab{b}}.
\newblock \href {https://doi.org/10.18653/v1/N19-1086} {Sentence embedding alignment for lifelong relation extraction}.
\newblock In \emph{Proceedings of the 2019 Conference of the North {A}merican Chapter of the Association for Computational Linguistics: Human Language Technologies, Volume 1 (Long and Short Papers)}, pages 796--806, Minneapolis, Minnesota. Association for Computational Linguistics.

\bibitem[{Welleck et~al.(2019)Welleck, Weston, Szlam, and Cho}]{welleck-etal-2019-dialogue}
Sean Welleck, Jason Weston, Arthur Szlam, and Kyunghyun Cho. 2019.
\newblock \href {https://doi.org/10.18653/v1/P19-1363} {Dialogue natural language inference}.
\newblock In \emph{Proceedings of the 57th Annual Meeting of the Association for Computational Linguistics}, pages 3731--3741, Florence, Italy. Association for Computational Linguistics.

\bibitem[{Williams et~al.(2018)Williams, Nangia, and Bowman}]{williams-etal-2018-broad}
Adina Williams, Nikita Nangia, and Samuel Bowman. 2018.
\newblock \href {https://doi.org/10.18653/v1/N18-1101} {A broad-coverage challenge corpus for sentence understanding through inference}.
\newblock In \emph{Proceedings of the 2018 Conference of the North {A}merican Chapter of the Association for Computational Linguistics: Human Language Technologies, Volume 1 (Long Papers)}, pages 1112--1122, New Orleans, Louisiana. Association for Computational Linguistics.

\bibitem[{Wu et~al.(2022)Wu, Caccia, Li, Li, Qi, and Haffari}]{wu2022pretrained}
Tongtong Wu, Massimo Caccia, Zhuang Li, Yuan-Fang Li, Guilin Qi, and Gholamreza Haffari. 2022.
\newblock Pretrained language model in continual learning: A comparative study.
\newblock In \emph{International Conference on Learning Representations}.

\bibitem[{Xu et~al.(2020)Xu, Zhang, Mao, Wang, Xie, and Zhang}]{xu-etal-2020-curriculum}
Benfeng Xu, Licheng Zhang, Zhendong Mao, Quan Wang, Hongtao Xie, and Yongdong Zhang. 2020.
\newblock \href {https://doi.org/10.18653/v1/2020.acl-main.542} {Curriculum learning for natural language understanding}.
\newblock In \emph{Proceedings of the 58th Annual Meeting of the Association for Computational Linguistics}, pages 6095--6104, Online. Association for Computational Linguistics.

\bibitem[{Yanaka et~al.(2020)Yanaka, Mineshima, Bekki, and Inui}]{yanaka-etal-2020-neural}
Hitomi Yanaka, Koji Mineshima, Daisuke Bekki, and Kentaro Inui. 2020.
\newblock \href {https://doi.org/10.18653/v1/2020.acl-main.543} {Do neural models learn systematicity of monotonicity inference in natural language?}
\newblock In \emph{Proceedings of the 58th Annual Meeting of the Association for Computational Linguistics}, pages 6105--6117, Online. Association for Computational Linguistics.

\bibitem[{Yanaka et~al.(2021)Yanaka, Mineshima, and Inui}]{yanaka-etal-2021-exploring}
Hitomi Yanaka, Koji Mineshima, and Kentaro Inui. 2021.
\newblock \href {https://doi.org/10.18653/v1/2021.eacl-main.78} {Exploring transitivity in neural {NLI} models through veridicality}.
\newblock In \emph{Proceedings of the 16th Conference of the European Chapter of the Association for Computational Linguistics: Main Volume}, pages 920--934, Online. Association for Computational Linguistics.

\bibitem[{Zhang et~al.(2018)Zhang, Dinan, Urbanek, Szlam, Kiela, and Weston}]{zhang-etal-2018-personalizing}
Saizheng Zhang, Emily Dinan, Jack Urbanek, Arthur Szlam, Douwe Kiela, and Jason Weston. 2018.
\newblock \href {https://doi.org/10.18653/v1/P18-1205} {Personalizing dialogue agents: {I} have a dog, do you have pets too?}
\newblock In \emph{Proceedings of the 56th Annual Meeting of the Association for Computational Linguistics (Volume 1: Long Papers)}, pages 2204--2213, Melbourne, Australia. Association for Computational Linguistics.

\end{thebibliography}
\bibliographystyle{acl_natbib}


\end{document}